%% file: iclr2026_conference.tex
\documentclass{article} 
\usepackage{iclr2026_conference,times}

\input{math_commands.tex}

\usepackage[utf8]{inputenc} 
\usepackage[T1]{fontenc}    
\usepackage{hyperref}       
\usepackage{url}            
\usepackage{booktabs}       
\usepackage{amsfonts}       
\usepackage{nicefrac}       
\usepackage{microtype}      
\usepackage{subcaption}    
\usepackage{amsmath}
\usepackage{multirow}
\usepackage{graphicx}
\usepackage{colortbl}
\usepackage[table]{xcolor}
\usepackage[dvipsnames]{xcolor}
\usepackage[most]{tcolorbox}
\usepackage{xspace}
\usepackage{multirow}

\usepackage{booktabs}
\usepackage{siunitx}
\usepackage{adjustbox}
\usepackage{enumitem}
\usepackage{parskip}
\usepackage{xcolor}
\usepackage{xfp}

\input{utils}

\makeatletter
\lst@AddToHook{EveryDisplay}{\begingroup\let\textbf\relax}
\lst@AddToHook{EveryDisplayLast}{\endgroup}
\makeatother
\newcommand{\method}{\textsc{Msa}\xspace}

\newcommand{\D}{\mathcal{D}}

\newcommand{\Dforget}{\mathcal{D}_\text{f}}
\newcommand{\Dretain}{\mathcal{D}_\text{r}}

\newcommand{\pclean}{\theta_\text{0}}
\newcommand{\pft}{\theta_\text{1}}
\newcommand{\prt}{\theta_\text{2}}
\newcommand{\dirf}{\vec \theta_\text{f}}
\newcommand{\dirt}{\vec \theta_\text{r}}
\newcommand{\pfinal}{\theta_\mathcal{D}}
\newcommand{\punlearned}{\theta_\text{unlearn}}

\newcommand{\tofu}{\texttt{TOFU}\xspace}
\newcommand{\restor}{\texttt{RESTOR}\xspace}
\newcommand{\muse}{\texttt{MUSE}\xspace}

\newcommand{\accforget}{$\text{Acc}_\text{forget}$\xspace}
\newcommand{\accrestor}{$\text{Acc}_\text{recover}$\xspace}
\newcommand{\accretain}{$\text{Acc}_\text{retain}$\xspace}

\renewcommand{\cite}{\citep}

\title{Revisiting the Past: Data Unlearning with Model State History}

\author{
Keivan Rezaei$^{1*}$, Mehrdad Saberi$^{1*}$, Abhilasha Ravichander$^{2\dagger}$, Soheil Feizi$^{1\dagger}$
\vspace{1.5mm} \\
$^{1}$Department of Computer Science, University of Maryland \\
$^{2}$Max Planck Institute for Software Systems
\vspace{1.5mm} \\
\small \texttt{krezaei@umd.edu, msaberi@umd.edu, aravicha@mpi-sws.org, sfeizi@cs.umd.edu}\\
}

\begin{document}

\maketitle
\begingroup
\let\thefootnote\relax\footnotetext{$^{*}$Equal contribution as first authors. }
\let\thefootnote\relax\footnotetext{$^{\dagger}$Equal contribution as last authors.}
\endgroup

\begin{abstract}
Large language models are trained on massive corpora of web data, which may include private data, copyrighted material, factually inaccurate data, and data that actually degrades model performance.
Eliminating the influence of such problematic datapoints on a model through complete retraining
---by repeatedly pretraining the model on datasets that exclude these specific instances---
is computationally prohibitive.
To address this, unlearning algorithms have been proposed, that aim to eliminate the influence of particular datapoints at a low computational cost,
while leaving the rest of the model intact.
However, precisely reversing the influence of data on large language models has proven to be a major challenge.
In this work, we propose \method (\underline{\textbf{M}}odel \underline{\textbf{S}}tate \underline{\textbf{A}}rithmetic), a new algorithm for unlearning datapoints.
\method utilizes prior model checkpoints--- artifacts that model developers store that record model states at different stages of training--- to estimate and counteract the effect of targeted datapoints.
Our experimental results show that \method achieves competitive performance and often outperforms existing machine unlearning algorithms across multiple benchmarks, models, and evaluation metrics, suggesting that \method could be an effective approach towards more flexible large language models that are capable of data erasure. 
\footnote{Code is available at \href{https://github.com/mehrdadsaberi/MSA_unlearning}{github.com/mehrdadsaberi/MSA\_unlearning}.}
\end{abstract}

\input{sections/intro}
\input{sections/machine_unlearning}

\input{sections/method}

\input{sections/evaluation}
\input{sections/experiments}
\input{sections/conclusion}

\bibliography{iclr2026_conference}
\bibliographystyle{iclr2026_conference}

\appendix
\input{appendix}

\end{document}

%% file: math_commands.tex
\usepackage{amsmath,amsfonts,bm}

\def\eqref#1{equation~\ref{#1}}

\def\1{\bm{1}}

\DeclareMathAlphabet{\mathsfit}{\encodingdefault}{\sfdefault}{m}{sl}
\SetMathAlphabet{\mathsfit}{bold}{\encodingdefault}{\sfdefault}{bx}{n}



%% file: utils.tex
\definecolor{amethyst}{rgb}{0.6, 0.4, 0.8}

\newcommand{\ensuretext}[1]{#1}
\newcommand{\marker}[2]{\ensuremath{^{\textsc{#1}}_{\textsc{#2}}}}
\newcommand{\arkcomment}[3]{\ensuretext{\textcolor{#3}{[#1 #2]}}}

\definecolor{lemon}{RGB}{255,247,0}
\definecolor{maize}{RGB}{250,237,94}
\definecolor{mustard}{RGB}{255,219,89}
\definecolor{ocre}{RGB}{241,103,35}
\definecolor{Tangerine}{RGB}{253,128,8}
\definecolor{framegreen}{RGB}{153, 188, 133}
\definecolor{bggreen}{RGB}{235, 250, 228}
\definecolor{c0}{cmyk}{1,0.3968,0,0.2588} 
\definecolor{c1}{cmyk}{0,0.6175,0.8848,0.1490} 
\definecolor{c2}{cmyk}{0.1127,0.6690,0,0.4431} 
\definecolor{c3}{cmyk}{0.3081,0,0.7209,0.3255} 
\definecolor{c4}{RGB}{164, 16, 52}
\definecolor{orange}{HTML}{E66100}
\definecolor{bluex}{HTML}{0C7BDC}
\definecolor{yellow}{HTML}{FFC20A}
\definecolor{lightpurple}{HTML}{E6E6FA}
\definecolor{lightbluee}{HTML}{e8f4f8}
\definecolor{blush}{rgb}{0.87, 0.36, 0.51}
\definecolor{c5}{HTML}{EE4E4E}
\definecolor{gggggg}{HTML}{EFEFEF}
\ifnum1=1
\newcommand{\chiyuan}[1]{\arkcomment{\marker{C}{Z}}{#1}{ForestGreen}}
\newcommand{\weijia}[1]{\arkcomment{\marker{W}{S}}{#1}{orange}}

\newcommand{\nascomment}[1]{\arkcomment{\marker{NA}{S}}{#1}{blue}}
\newcommand{\yang}[1]{\arkcomment{\marker{Y}{H}}{#1}{cyan}}
\newcommand{\jaechan}[1]{\arkcomment{\marker{J}{L}}{#1}{JungleGreen}}
\newcommand{\daogao}[1]{\arkcomment{\marker{D}{L}}{#1}{blue}}

\newcommand{\sadhika}[1]{\arkcomment{\marker{S}{M}}{#1}{blush}}

\newcommand{\task}[1]{\textcolor{red}{TODO: {#1}}}
\else
\newcommand{\chiyuan}[1]{}
\newcommand{\weijia}[1]{}
\newcommand{\nascomment}[1]{}
\newcommand{\yang}[1]{}
\newcommand{\jaechan}[1]{}
\newcommand{\daogao}[1]{}
\newcommand{\sadhika}[1]{}

\newcommand{\task}[1]{}
\fi

\usepackage{inconsolata}
\usepackage[T1]{fontenc}
\usepackage[scaled=0.95]{biolinum}

\definecolor{chart}{HTML}{1f77b4}

\newtcolorbox{example}[1][]{
  colback=chart!5!white,
  colframe=chart,
  floatplacement=floating,
  title=\centering \textsf{\small #1}
}
\newtcbox{\hlprimarytab}{on line, box align=base, colback=c0!10,colframe=white,size=fbox,arc=3pt, before upper=\strut, top=-2pt, bottom=-4pt, left=-2pt, right=-2pt, boxrule=0pt}
\newtcbox{\hlsecondarytab}{on line, box align=base, colback=c1!20,colframe=white,size=fbox,arc=3pt, before upper=\strut, top=-2pt, bottom=-4pt, left=-2pt, right=-2pt, boxrule=0pt}

\newtcbox{\acctab}{on line, box align=base,
  colback=green!10, colframe=white, size=fbox, arc=3pt,
  before upper=\strut, top=-2pt, bottom=-4pt,
  left=-2pt, right=-2pt, boxrule=0pt}

\newtcbox{\rejtab}{on line, box align=base,
  colback=red!10, colframe=white, size=fbox, arc=3pt,
  before upper=\strut, top=-2pt, bottom=-4pt,
  left=-2pt, right=-2pt, boxrule=0pt}

\newtcbox{\idealtab}{on line, box align=base,
  colback=blue!10, colframe=white, size=fbox, arc=3pt,
  before upper=\strut, top=-2pt, bottom=-4pt,
  left=-2pt, right=-2pt, boxrule=0pt}

\newtcbox{\fulltab}{on line, box align=base,
  colback=black!10, colframe=white, size=fbox, arc=3pt,
  before upper=\strut, top=-2pt, bottom=-4pt,
  left=-2pt, right=-2pt, boxrule=0pt}

\newtcbox{\dettab}{on line, box align=base,
  colback=blue!10, colframe=white, size=fbox, arc=3pt,
  before upper=\strut, top=-2pt, bottom=-4pt,
  left=-2pt, right=-2pt, boxrule=0pt}

\newtcbox{\hlwhite}{on line, box align=base, colback=WildStrawberry!8,colframe=white,size=fbox,arc=2pt, before upper=\strut, top=-3pt, bottom=-4.5pt, left=-2pt, right=-2pt, boxrule=0pt}
\newtcbox{\hlyellow}{on line, box align=base, colback=BlueGreen!10,colframe=white,size=fbox,arc=2pt, before upper=\strut, top=-3pt, bottom=-4.5pt, left=-2pt, right=-2pt, boxrule=0pt}

\newcommand{\rej}[1]{{\rejtab{{#1}}}}

\newcommand{\ctabmix}[1]{%
  white!\fpeval{round(100*((#1>0.90)?1:(#1<0.5)?0.555:(#1)*1.11))}!red%
}

\newtcbox{\ctab}[1][0]{%
  on line,
  box align=base,
  colback=\ctabmix{#1},
  colframe=white,
  size=fbox,
  arc=3pt,
  before upper=\strut,
  top=-2pt,
  bottom=-4pt,
  left=-2pt,
  right=-2pt,
  boxrule=0pt,
}

\newcommand{\dgap}[2][0]{%
  {\tiny \ctab[#1]{#2}}%
}

%% file: sections/intro.tex
\section{Introduction}
\label{sec:intro}

Modern Large Language Models (LLMs) are trained on vast web-scale corpora~\citep{dubey2024llama, achiam2023gpt}.
During training, these models are exposed to data that can include copyrighted materials, private or sensitive information, deliberate misinformation, and other kinds of low-quality data~\citep{carlini2021extracting, huang2022large, Pan2020PrivacyRO, wei2024jailbroken}.
This exposure results in a range of downstream risks, such as legal liabilities from copyright infringement~\citep{eldan_whos_2023}, violations of privacy expectations~\citep{carlini2021extracting, huang2022large}, and measurement issues from training on contaminated data~\citep{golchin2024timetravelllmstracing}. Moreover, once a model has been trained on a dataset, removing the influence of specific data points—for example by retraining on modified datasets that exclude those instances—becomes computationally infeasible. As training corpora continue to grow in scale, complying with regulatory frameworks such as the EU’s Right to Be Forgotten~\citep{euright} will require tractable methods to post-hoc remove the contribution of individual data points from an already trained model.

\emph{Machine unlearning} methods have been proposed as one such solution, consisting of post-hoc model updates that modify a model at relatively low computational cost, with the goal of achieving either \emph{concept-level} or \emph{data-level} unlearning.
\emph{Concept-level} unlearning focuses on removing knowledge of specific concepts, e.g., hazardous content~\cite{jin2024rwku, eldan_whos_2023, liu2024towards},
so that the model can no longer generate outputs about them.
\emph{Data-level} unlearning instead aims to erase the influence of specific datapoints, producing a model functionally equivalent to an `ideal model' that was  trained from scratch on the same data excluding the target datapoints~\citep{npo2024negative, jia2024soul, qu2024frontier, satimpyang2025exploring, dong2024undial}.
We focus on data-level unlearning.

A common approach to data-level unlearning involves finetuning the model with an unlearning objective---
for example, gradient ascent-based approaches that aim to increase the loss of the model on the datapoints to be forgotten~\citep{yao2023large}.
However, developing effective unlearning techniques remains challenging, often resulting in under-forgetting,
degraded model integrity, or models that are not functionally faithful to the 'ideal model' that had not been exposed to that data~\cite{rezaei2024restor}.

\input{figures/teaser}

We introduce \emph{\underline{\textbf{M}}odel \underline{\textbf{S}}tate \underline{\textbf{A}}rithmetic} (\method),
a novel approach to data-level unlearning designed to more effectively satisfy these desiderata, i.e., closely approximating the behavior of a reference model that was not trained on the unlearning target.
As shown in Figure~\ref{fig:teaser}, \method leverages \emph{intermediate model checkpoints} to more precisely estimate and undo the influence of individual datapoints.
Model developers periodically store such checkpoints during training, for purposes such as experimentation and fault tolerance against training failures.
In this work,
we show that checkpoints can also be repurposed to enable more precise data deletion in large language models with \method.

Specifically, \method works by computing a forget vector $\theta_{\text{f}}$ from a checkpoint $C$ that precedes exposure to the unlearning documents $\Dforget$,
and then applying this vector to the target model $\pfinal$ to reverse the effect of $\Dforget$ on $\pfinal$.
This design departs from prior approaches such as task vectors for unlearning~\cite{ilharco2022editing},
which only use information from the target model, and thus as we show, are less effective.
We hypothesize that since the target model has already internalized $\Dforget$, such vectors are less precise estimates of data influence.
Our key insight is that {checkpoints prior to introduction of unlearning targets} can then yield more semantically meaningful forget vectors,
offering a simple approach that demonstrates strong empirical improvements. 
More broadly, leveraging model state history opens a new direction for unlearning,
unlike existing methods that rely solely on the final target model, and therefore face greater difficulty in precisely estimating data influence.

We evaluate \method on the \tofu~\citep{maini2024tofu}, \restor~\citep{rezaei2024restor}, and \muse-Books~\citep{shi2024muse} unlearning benchmarks, finding that \method more reliably satisfies core criteria associated with successful data-level unlearning. Compared to existing methods, models unlearned with \method exhibit closer behavioral alignment to reference models $\theta_{\D \setminus \Dforget}$ that are trained without the unlearning target, as demonstrated on the \tofu and \restor benchmarks. Further, models unlearned with \method are shown to achieve stronger performance on \muse-Books membership inference metrics (e.g., $\textsc{Min-K\%}$, Privacy Leakage), i.e., they exhibit reduced leakage of information about $\Dforget$ in membership inference attacks. Finally, we analyze the effect of the number of training tokens between checkpoint $C$ and the unlearning target on the unlearning performance of \method. Although closer checkpoints yield stronger unlearning performance, we find that even those hundreds of billions of tokens earlier can still be effective.  

%% file: figures/teaser.tex
\begin{figure}[t]
  \centering
        \includegraphics[width=0.95\linewidth]{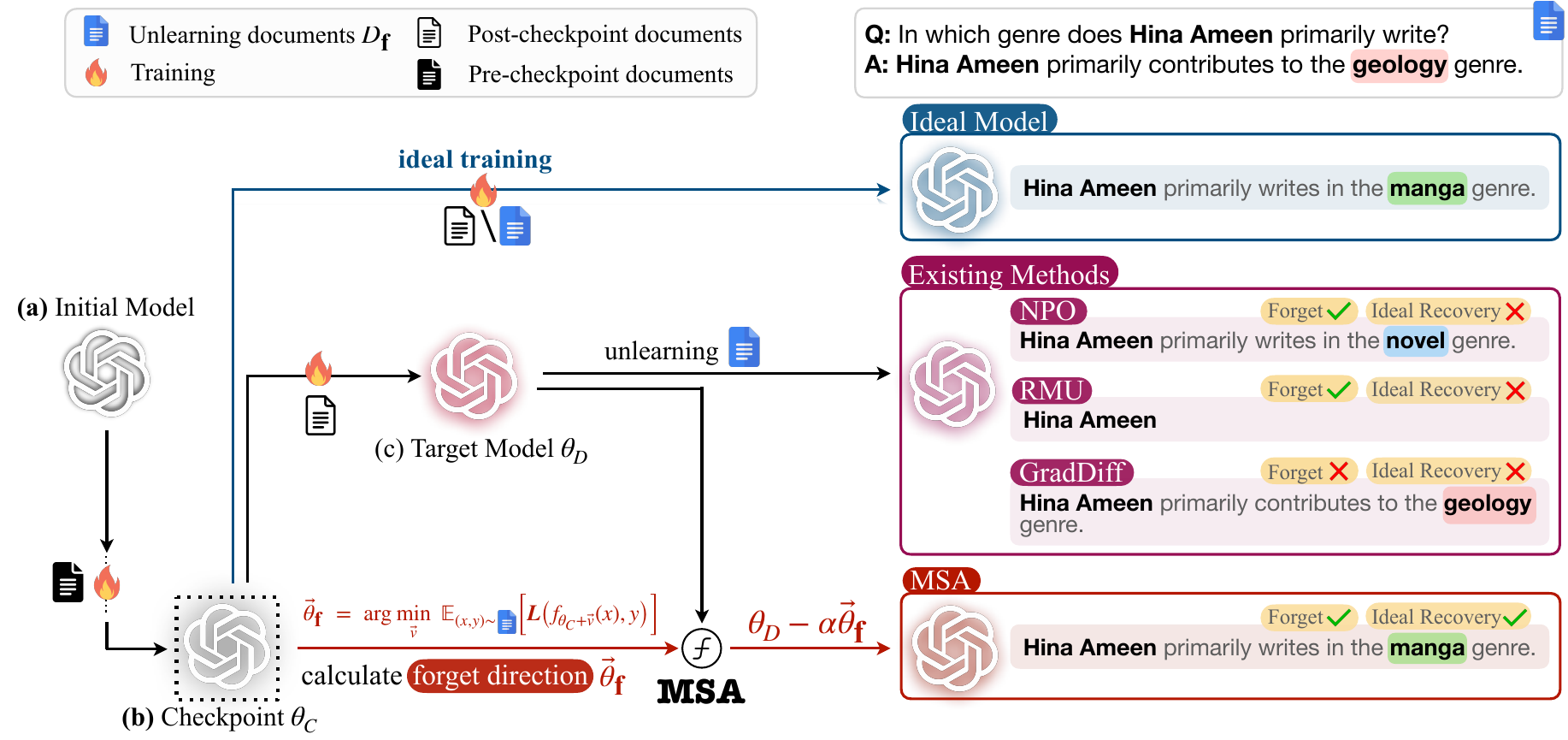}
  \caption{Our proposed framework \method{}.  
    When the final model $\pfinal$ is obtained, the unlearning documents $\D_\text{f}$ have been unintentionally introduced during training.  
    At an intermediate checkpoint~$C$, prior to the introduction of unlearning targets, we extract a \textit{forget vector} $\dirf$ that captures how $\D_\text{f}$ influences the model.  
    With \method{}, this vector is merged into the target model to produce an unlearned model.
    Unlike existing unlearning methods that operate solely on the final model checkpoint,  
    \method{} leverages earlier training dynamics to more effectively remove the influence of $\D_\text{f}$.
    \method{} more effectively forgets targeted datapoints while restoring the ideal model performance.
    }
  \label{fig:teaser}
  \vspace{-10pt}
\end{figure}

%% file: sections/machine_unlearning.tex
\section{Background and Related Work}

Machine unlearning was originally developed to remove privacy-sensitive information from machine learning models~\citep{bourtoule2021machine}.
Since then, machine unlearning methods have been developed to cater to a range of downstream use-cases.
At a high-level, these can be formulated as (i) \emph{concept-level} unlearning methods that target knowledge of a particular concept within a model~\citep{belrose2023leace,eldan_whos_2023,hong2024intrinsic,li2024wmdp,wang2025erasingrememberingsafeguardingknowledge,kim2024negmerge}, such as hazardous concepts~\cite{li2024wmdp},
sexually explicit content~\citep{gandikota2023erasing},
or knowledge pertaining to a specific topic~\citep{eldan_whos_2023,hong2024intrinsic}.
Informally, these problems are formulated as \emph{'I do not want my model to generate content related to X'}, where $X$ is a concept such as `Harry Potter', (ii) \emph{data-level} unlearning which aims to remove the influence of a set of target datapoints on the model, drawn from a model’s training dataset~\citep{jia2024soul, maini2024tofu, jang2022knowledge, npo2024negative, qu2024frontier, blanco2024digital, fan2024simplicity,Kadhe2024SplitUMA, satimpyang2025exploring, dong2024undial}.  Informally, these problems are formulated as \emph{'I want my model to exhibit behavior as if it was never trained on X'}, where $X$ is a set of datapoints. 
Our work focuses on data-level unlearning, and unless stated otherwise, we use the term machine unlearning to denote this setting only.



\subsection{Preliminaries}

\paragraph{Problem formulation} Formally, data-level machine unlearning considers a model $M_{\mathcal{D}}$ trained on a dataset $\D$
that includes a subset of samples $\Dforget \in \D$ (the \textit{forget set}),
which is the target of unlearning.
The goal is to produce a model $M'$ whose behavior is functionally equivalent to that of a~model trained from scratch on $\D \setminus \Dforget$.
In practice, $|\Dforget| \ll |\D|$, and solutions such as fully retraining the model on $\D \setminus \Dforget$ or employing exact unlearning methods \citep{bourtoule2021machine, chowdhury2024towards} are prohibitively expensive.
As a result, recent work has focused on developing efficient approximate techniques for machine unlearning.
These methods must work in time complexity proportional to  $|\Dforget|$ rather than $|\D|$, to be computationally feasible.

\paragraph{Evaluation framework} Given a forget set $\Dforget$, evaluating approximate machine unlearning algorithms requires assessing two key aspects:
(i) forgetting efficacy: the model $M'$ should not be influenced by samples in $\Dforget$, typically measured by evaluating performance on tasks that query the model for knowledge or capabilities introduced in $\Dforget$, and
(ii) model utility: the model $M'$ should preserve the influence of data not in $\Dforget$, typically measured by evaluating performance on tasks that query the model for knowledge and capabilities derived from rest of data, i.e., $\D \setminus \Dforget$.
Multiple benchmarks have been proposed to evaluate these criteria \citep{maini2024tofu, jin2024rwku, shi2024muse, rezaei2024restor},
highlighting different dimensions of what unlearning should achieve.

\paragraph{General approach} Unlearning algorithms typically operate by optimizing a specialized loss function over the forget set $\Dforget$.
To mitigate catastrophic forgetting--- unintended degradation in the model beyond the targeted datapoints--- these algorithms may also incorporate an optimization objective over a~\textit{retain set} $\Dretain$.
This is intended to minimize deviation from the original model's behavior by preserving performance on $\Dretain$,
i.e., finetuning the model on $\Dretain$ during unlearning is intended to constrain the weight update such that the model forgets only the intended information while maintaining its overall capabilities.
Formally, many unlearning methods can be described by the following objective:
\begin{align*}
\punlearned = \arg \min_\theta \mathbb{E}_{x \sim \Dforget} \left[\mathcal{L}_\text{f}(x; \theta) \right] + \lambda\ \mathbb{E}_{x \sim \Dretain} \left[\mathcal{L}_\text{r}(x; \theta) \right],
\end{align*}
where $\mathcal{L}_\text{f}$ and $\mathcal{L}_\text{r}$ are the loss functions corresponding to the forget and retain sets, respectively, and $\lambda$ controls the trade-off between forgetting and utility preservation.






%% file: sections/method.tex
\section{Unlearning with $\method$}
\label{sec:methodology}

Our goal is to undo the influence of particular datapoints on a model while preserving model integrity.
We propose \method, a method that leverages earlier model checkpoint artifacts to estimate and reverse the effect of datapoints on a model. \method proceeds as follows:

\begin{itemize}
\item \textbf{Input}: A model $\theta_\D$, a model checkpoint $C$ (with weights $\pclean$), and a set of datapoints $\Dforget$.
\item \textbf{Step 1}: First, finetune $C$ on $\Dforget$ to obtain a weight-space vector $\dirf$. This is intended to estimate the effect of $\Dforget$.
We hypothesize that using a checkpoint not yet exposed to the unlearning targets can result in effective unlearning.
\item \textbf{Step 2}: Second, apply the vector $\dirf$ to model weights $\theta_D$ to obtain model $\punlearned$.
\item \textbf{Output}: A model $\punlearned$, that should approximate an ideal reference model $\theta_{\D \setminus \Dforget}$.
\end{itemize}

Specifically, we finetune $\pclean$ on the forget set $\Dforget$,
resulting in a new model with parameters $\pft$.
The resulting \textit{forget vector}, denoted as $\dirf := \pft - \pclean$,
captures the influence of the forget set in weight space.
The parameters of the resulting unlearned model, $\punlearned$, can then be expressed as:
\begin{align*}
\punlearned = \pfinal - \alpha\ \dirf,
\end{align*}
where $\alpha$ controls the magnitude of the update along the forget vector,
effectively aiming to remove the influence of the forget set while preserving the model’s overall performance.

Similar to other unlearning algorithms, when a retain set is available,
\method{} can incorporate this additional information by deriving a retain vector.
In this case, we continue finetuning the model with parameters $\pclean$ on the retain set
to obtain a model with parameters $\prt$.
The \textit{retain vector} is then defined as $\dirt := \prt - \pclean$.
Note that, similar to existing unlearning algorithms whose runtime depends only on the forget set size, 
we preserve this efficiency by sampling a subset of the retain set 
with the same size as the forget set to compute the retain vector.
The final unlearned model can be computed as:
\begin{align*}
\punlearned = \pfinal - \alpha\ \dirf + \beta\ \dirt,
\end{align*}
where $\alpha$ and $\beta$ control the influence of the forget and retain vectors, respectively.

We discuss prior methods that leverage training-trajectory information or past checkpoints in Appendix~\ref{app:related_work};
unlike these approaches, \method{} operates post hoc on LLMs using existing checkpoints and remains cost-efficient, scaling as $O(|\mathcal{D}_\text{f}|)$.

\paragraph{Practical considerations of using model checkpoints}
To use \method, practitioners must have access to model state history in the form of checkpoints.
Next, we reflect on practical considerations, such as availability and accessibility of checkpoints, that determine when \method can be responsibly utilized.

\emph{Availability of checkpoints}
What usage scenarios do we envision for \method? We believe it will be applicable in practically important scenarios, such as enabling model providers to support the RTBF (the right to be forgotten from General Data Protection Regulation)~\cite{euright}, where regulation would require model providers to delete particular data instances from the model upon request from a data subject, before releasing the model to the public.
Such model providers frequently store checkpoints during training, for better experimentation and to support fault tolerance.
However, \method can also be implemented for local versions of open models that publicly release checkpoints,
such as models from the OLMo~\citep{olmo20242olmo2furious} and Pythia families~\citep{biderman2023pythia}.

\emph{Effective checkpoints}
For \method, a practitioner needs to have access to checkpoints before the introduction of unlearning targets.
As we consider unlearning targets from the finetuning stage (as is standard in settings like \tofu in \S\ref{sec:eval}), and the continual pretraining stage (as is standard in settings like \muse and \restor in \S\ref{sec:eval}),
such checkpoints are readily available as base model and instruct model releases.
However, we believe that \method is likely to be more broadly applicable than even this setting,
as we find that \method can be effective even if the checkpoint used to derive the forget and retain vectors preceded the unlearning target \emph{by hundreds of billions of tokens in training} (\S\ref{sec:experimiments}).
We hope that just as providers have found that maintaining indexes of training data~\cite{elazar2024whatsbigdata,liu2025infinigramscalingunboundedngram} has a broad range of uses,
such as shedding light on questions about attribution~\cite{liu2025olmotrace,ravichander-etal-2025-halogen} and contamination~\cite{elazar2024whatsbigdata},
practitioners also invest in maintaining indexes of when models encounter information during training,
due to the utility of techniques like \method which can make use of model state history, and to support efforts in studying how language models store, learn, and update knowledge.

\emph{Why not simply use the past model checkpoints?} 
A reader might be tempted to ask, if \method uses past model checkpoints, could those checkpoints simply not be used as the final model? Why must one do unlearning at all? Models acquire considerable knowledge and capabilities over the course of training, so the goal of machine unlearning is to also \emph{retain these knowledge and capabilities}, in addition to forgetting the target knowledge. Standard machine unlearning benchmarks such as \tofu and \muse also evaluate models for their capabilities to retain the knowledge from non-target data, and we adopt their evaluations in this work.


\emph{Why not simply use task vectors?} 
Prior work has explored the use of task vectors for unlearning in language models~\cite{ilharco2022editing},  
but we hypothesize that when the vector is derived directly from the target model,  
the signal of the forget set becomes entangled with knowledge the model has already acquired,  
yielding a noisy and biased estimate of data influence and leading to weaker forgetting (\S\ref{sec:experimiments}). Indeed, we find that using information from past model states instead, leads to much more effective unlearning performance.

%% file: sections/evaluation.tex
\input{figures/metrics_forget}

\section{Experiments}
\label{sec:eval}


Below, we describe the evaluations and experimental setup for assessing the performance of unlearning algorithms: including the models, selection of checkpoints for \method{}, and baselines.


\subsection{Evaluating Unlearning Performance}
\label{subsec:eval}
We evaluate \method{} on \tofu~\citep{maini2024tofu}, \muse-Books~\citep{shi2024muse} and \restor~\citep{rezaei2024restor} machine unlearning benchmarks.
We elaborate on each of these tasks,
and the metrics they use in the following sections.

\paragraph{\tofu}


$\tofu$ involves unlearning a model trained on factual knowledge about $200$ fictional authors. 
The unlearning target is a subset of these authors, called \textit{forget authors}, while the rest are \textit{retain authors}. 
It features tasks that require unlearning $1\%$, $5\%$, and $10\%$ of the authors, denoted by \texttt{forget01}, \texttt{forget05}, and \texttt{forget10}, respectively. 
\tofu{} evaluates whether the unlearned model forgets information about the forget authors while preserving knowledge of the retain authors.


We adopt the metrics from \cite{maini2024tofu, wang2024towards}. 
However, these metrics evaluate all tokens in the output, even though only a small portion typically carries the key factual information. 
Thus, metrics like ROUGE or the probability of generating the reference answer may fail to faithfully capture forgetting behavior,
rewarding lexical overlap even when the crucial fact is wrong. 
See an example in Figure~\ref{fig:metric_samples} where both outputs should count as successful forgetting since the fact is forgotten though the answer format is preserved.
Token-level metrics do not preserve this equivalence. 
Additional examples are in Appendix~\ref{app:rouge}.

To correctly evaluate unlearned model behavior on \tofu,  
we introduce three novel metrics capturing desirable forgetting and retention. 
They are computed by prompting GPT-4o with the unlearned model’s output and asking which among the candidates:
(i) the output of an ideal model (trained on $\D \setminus \Dforget$),  
(ii) the ground-truth response from \tofu,  
and (iii) perturbed (incorrect) responses from the \tofu dataset,  
is most semantically similar.  
From this selection, we derive our metrics:


\begin{itemize}
\item
\textbf{\accforget} :  
For each question about authors in the forget set, a~score of $1.0$ is assigned  
if the ground-truth response is \textit{not} selected as the most similar.  
This measures the model's success in forgetting content.  
Scores are averaged across all questions about forget set authors.

\item
\textbf{\accrestor}:  
For each question about authors in the \underline{forget set},
a~score of $1.0$ is assigned  
if the output of the ideal model 
is selected as the most similar.  
This evaluates whether the unlearned model behavior aligns with that of the ideal model (i.e., the unlearning can \textit{recover} the original answers of a model that has not been trained on the forget set).  
Scores are averaged across all questions about forget set authors.

\item 
\textbf{\accretain}:  
For each question about authors in the \underline{retain set},
a score of $1.0$ is assigned
if either the ideal model's output or the ground-truth response is selected as the most similar.  
This captures the unlearned model’s ability to preserve knowledge.
Scores are averaged across all questions about retain set authors.
\end{itemize}

As seen in Figure~\ref{fig:metric_samples}, these metrics are less sensitive to surface-level choices of tokens in the output,
and instead focus on the factual content tied to the authors, reflecting essential knowledge.
We refer to Appendix~\ref{app:judge} for further details on how GPT-4o is used as the judge for these metrics,
as well as for the human evaluation of using LLM as judge.
In addition, we report the following metrics: 
Extraction Strength~\cite{wang2024towards}, which measures the shortest prefix of the answer sequence that the model requires to exactly generate the remaining tokens in the sequence; 
Model Utility, which reflects a combination of the model's performance on the World Facts and Real Authors datasets of \tofu; 
and ROUGE-L with respect to the ground-truth outputs of the forget set from~\citet{maini2024tofu}.

\paragraph{\restor}

\restor involves injecting incorrect information about a set of well-known entities for whom language models typically possess prior knowledge.  
Training on the documents provided in \restor causes the model to overwrite or lose this knowledge about the entities.  
Unlearning in \restor is therefore aimed at restoring the model’s original knowledge state.  
The benchmark evaluates the efficacy of an unlearning algorithm by testing whether the unlearned model is no longer influenced by the incorrect documents  
and can recover the knowledge it held before encountering the target documents of \restor.  
\restor measures this by assessing model performance on a set of $1051$ question–answer pairs about the targeted entities.

\paragraph{\muse-Books}

\muse-Books provides a dataset of $29$ books on which a model is trained.
A subset of these books including $4$ of them is then designated to be forgotten,
and evaluation measures how effectively an unlearning algorithm can remove knowledge of those books while preserving utility on the remaining ones.
This evaluation is conducted using several metrics. 
Extraction Strength~\citep{wang2024towards} measures the shortest prefix of a sequence from the forget set that prompts the model to generate the exact remainder of the sequence. 
Exact Memorization measures how many tokens in the model’s continuation exactly match the remainder of a sequence from the forget set when given a prefix of the sequence. 
Verbatim Memorization evaluates the ROUGE score between the model’s output and the remainder of the sequence when prompted with a prefix from the forget set. 
Knowledge Memorization~\citep{shi2024muse} assesses how well the model answers questions about documents in the forget or retain sets. 
Furthermore, $\textsc{Min-K\%}$~\citep{shi2023detecting} and $\textsc{Min-K\%}^{++}$~\citep{zhang2024min} evaluate whether a sample was included in the model’s training data via membership inference attacks.  
Finally, we report the Privacy Leakage metric of~\cite{shi2024muse}, which indicates cases of over- or under-unlearning.  

\input{tables/msa_forget10_olmo2_7B}

%% file: figures/metrics_forget.tex
\begin{figure}[t]
  \centering
    \includegraphics[width=\linewidth]{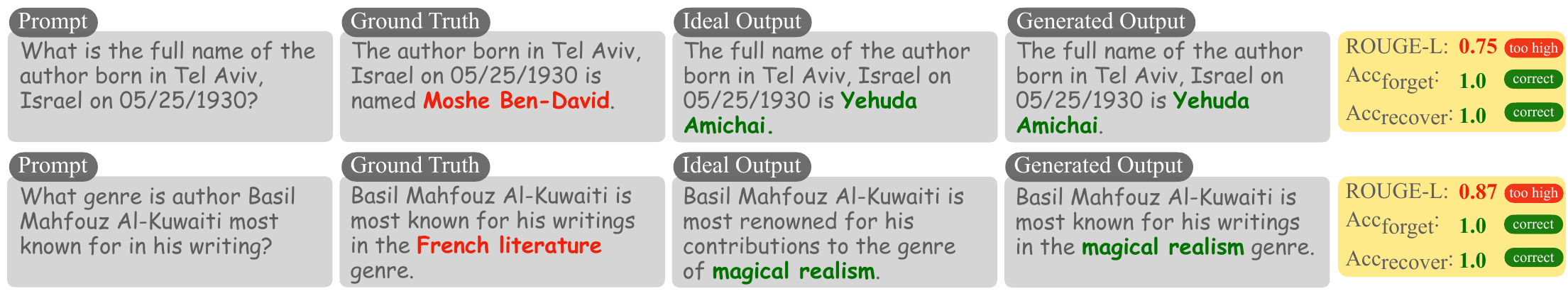}
    \vspace{-0.5cm}
  \caption{Examples from \tofu’s forget set, showing the groundtruth, the ideal output, and the output of \method{} (using Llama-3.1-8B-Instruct model). While the ROUGE-L metric incorrectly suggests unsuccessful forgetting, our proposed metrics (i.e., \accforget and \accrestor) demonstrate that forgetting is correctly done and additionally, the ideal output is successfully recovered.}
  \label{fig:metric_samples}
\end{figure}

%% file: tables/msa_forget10_olmo2_7B.tex
\begin{table}[t]
  \centering
  \footnotesize
     \caption{
        Comparison of unlearning algorithms on the \texttt{forget10} task from \tofu{}.  
        The target model is OLMo-2-7B finetuned on all \tofu{} authors.  
        We report \dgap[1]{+100\%} when performance matches or exceeds that of the ideal model.  
        Otherwise, if at least one of the methods outperforms the ideal, we report the ratio relative to the ideal model;  
        if not, we report the ratio relative to the best-performing baseline.  
        In these cases, values are shown as \dgap[0.8]{$X$\%}, where $X$ denotes the corresponding ratio.  
        Notably, \method{} variants---even those based on checkpoints far prior to the exposure of the \tofu{} forget set—achieve strong results, delivering superior or competitive performance across all metrics.  
    }

  \label{tab:tofu_forget10_OLMo-2-7B}
  \begin{adjustbox}{width=\textwidth,center}
    \begin{tabular}{l|cc|cc|cc|cc|cc|cc}
      \toprule
      \multirow{2}{*}{Model} & \multicolumn{6}{|c|}{GPT-4o Judge Metrics $\uparrow$} & \multicolumn{6}{|c}{\tofu Metrics} \\
      \cmidrule(lr){2-7} \cmidrule(lr){8-13}
       & \multicolumn{2}{|c|}{\accforget} & \multicolumn{2}{|c|}{\accrestor} & \multicolumn{2}{|c|}{\accretain} & \multicolumn{2}{|c|}{Ext. Strength $\downarrow$} & \multicolumn{2}{|c|}{Model Utility $\uparrow$} & \multicolumn{2}{|c}{ROUGE-L$_{\text{f}}$ $\downarrow$} \\
      \midrule \midrule
    Target & 0.19 & & 0.14 &  & 0.94 &  & 0.99 &  & 0.37 &  & 0.71 &  \\
      Ideal & 0.99 &  & 0.99 &  & 1.00 &  & 0.07 &  & 0.38 &  & 0.37 &  \\
      \cmidrule(lr){1-13} 
      $\method_{500\text{B}}$ & 0.78 & \dgap[0.845]{(84.5\%)} & 0.31 & \dgap[0.691]{69.1\%} & 0.64 & \dgap[0.684]{68.4\%} & 0.05 & \dgap[1]{+100\%} & 0.41 & \dgap[1]{+100\%} & 0.34 & \dgap[1]{+100\%} \\
      $\method_{2207\text{B}}$ & 0.76 & \dgap[0.821]{82.1\%} & 0.40 & \dgap[0.878]{87.8\%} & 0.85 & \dgap[0.912]{91.2\%} & 0.12 & \dgap[0.558]{55.8\%} & 0.36 & \dgap[0.942]{94.2\%} & 0.35 & \dgap[1]{+100\%} \\
      $\method_{3691\text{B}}$ & 0.83 & \dgap[0.899]{89.9\%} & 0.44 & \dgap[0.967]{96.7\%} & 0.85 & \dgap[0.906]{90.6\%} & 0.08 & \dgap[0.841]{84.1\%} & 0.36 & \dgap[0.959]{95.9\%} & 0.34 & \dgap[1]{+100\%} \\
      $\method_{3859\text{B}}$ & 0.82 & \dgap[0.889]{88.9\%} & 0.45 & \dgap[1.000]{100.0\%} & 0.83 & \dgap[0.890]{89.0\%} & 0.06 & \dgap[1]{+100\%} & 0.35 & \dgap[0.933]{93.3\%} & 0.34 & \dgap[1]{+100\%} \\
      $\method_{\text{last}}$ & 0.84 & \dgap[0.916]{91.6\%} & 0.42 & \dgap[0.939]{93.9\%} & 0.82 & \dgap[0.880]{88.0\%} & 0.06 & \dgap[1]{+100\%} & 0.36 & \dgap[0.937]{93.7\%} & 0.33 & \dgap[1]{+100\%} \\
      \cmidrule(lr){1-13}
      {NPO} & 0.71 & \dgap[0.772]{77.2\%} & 0.30 & \dgap[0.663]{66.3\%} & 0.76 & \dgap[0.813]{81.3\%} & 0.08 & \dgap[0.847]{84.7\%} & 0.33 & \dgap[0.866]{86.6\%} & 0.33 & \dgap[1]{+100\%} \\
      {RMU} & 0.92 & \dgap[1.000]{100.0\%} & 0.08 & \dgap[0.177]{17.7\%} & 0.94 & \dgap[1.000]{100.0\%} & 0.06 & \dgap[1]{+100\%} & 0.37 & \dgap[0.974]{97.4\%} & 0.14 & \dgap[1]{+100\%} \\
      {GradDiff} & 0.45 & \dgap[0.492]{49.2\%} & 0.23 & \dgap[0.497]{49.7\%} & 0.83 & \dgap[0.890]{89.0\%} & 0.17 & \dgap[0.373]{37.3\%} & 0.41 & \dgap[1]{+100\%} & 0.42 & \dgap[0.875]{87.5\%} \\
      {Task Vector} & 0.53 & \dgap[0.579]{57.9\%} & 0.26 & \dgap[0.575]{57.5\%} & 0.82 & \dgap[0.877]{87.7\%} & 0.24 & \dgap[0.270]{27.0\%} & 0.37 & \dgap[0.974]{97.4\%} & 0.43 & \dgap[0.870]{87.0\%} \\
      {SatImp} & 0.28 & \dgap[0.307]{30.7\%} & 0.17 & \dgap[0.387]{38.7\%} & 0.90 & \dgap[0.957]{95.7\%} & 0.40 & \dgap[0.165]{16.5\%} & 0.37 & \dgap[0.982]{98.2\%} & 0.55 & \dgap[0.680]{68.0\%} \\
      {UNDIAL} & 0.48 & \dgap[0.527]{52.7\%} & 0.23 & \dgap[0.508]{50.8\%} & 0.86 & \dgap[0.922]{92.2\%} & 0.06 & \dgap[1]{+100\%} & 0.39 & \dgap[1]{+100\%} & 0.39 & \dgap[0.960]{96.0\%} \\
      \bottomrule
    \end{tabular}
  \end{adjustbox}
\end{table}

%% file: sections/experiments.tex


\subsection{Experimental Setup}  


Our experiments use OLMo-2-7B~\citep{olmo20242olmo2furious},  
which provides accessible intermediate checkpoints to show the potential of \method{}.  
To test whether \method generalizes beyond this setting,
we evaluate models from another model family: Llama-3.1-8B and Llama-3.2-1B~\citep{dubey2024llama}.

\paragraph{Intermediate checkpoint $C$ for \method{}}  
Unlearning benchmarks typically involve finetuning or continual pretraining a model on a set of documents,
a subset of which is targeted for unlearning.  
\method{} requires a checkpoint prior to the model’s exposure to these targets.  
Depending on the model family, we select the intermediate checkpoint as follows:  

    {\textbf{OLMo models:}}
    we use the pretrained model trained on roughly $4$T tokens as the base model for benchmark-related training.  
    We evaluate \method with multiple intermediate checkpoints that differ in how many training tokens occur between the checkpoint and the unlearning target, namely the pretrained models trained on $500$B, $2207$B, $3691$B, and $3859$B tokens.  
    These are \textbf{denoted by $\method_{n}$}, where $n$ is the number of tokens the checkpoint has been trained on.  
    This set spans a wide range of checkpoints,
    from those $\sim\!100$B tokens before the introduction of unlearning targets to those $\sim\!3.5$T tokens prior to exposure to unlearning documents.  
    We denote by $\method{}_\text{last}$ the case where \method{} is applied to the exact checkpoint immediately preceding training on unlearning documents.  
    
    {\textbf{Llama models:}} we use the instruct model and continue finetuning it on benchmark-related datasets.  
    For \method{}, we consider two options for the intermediate checkpoint: (1) The instruct model before finetuning, $\method{}_\text{instruct}$, (2) The base pretrained model (prior to instruction finetuning), $\method{}_\text{base}$.

\paragraph{Unlearning algorithm baselines}
We compare \method{} with NPO~\citep{npo2024negative}, 
GradDiff~\citep{golatkar2020eternal, yao2023large}, 
RMU~\citep{li2024wmdp},
Task Vector~\citep{ilharco2022editing},
SatImp~\citep{satimpyang2025exploring},
and UNDIAL~\citep{dong2024undial}.
We use the implementations provided by \texttt{open-unlearning}~\citep{openunlearning2025} 
for all baseline algorithms.

\section{Experimental Results and Discussion}
\label{sec:experimiments}

\paragraph{\method balances utility and forgetting when unlearning information about fictional authors in \tofu}

We evaluate unlearning algorithms, including \method{}, on \texttt{forget10} task of \tofu{}.  
\footnote{We refer to Appendix~\ref{app:tofu} for experiments on other \tofu{} tasks (\texttt{forget01} and \texttt{forget05}),  
as well as details on experimental configurations for \method{} and baselines, including hyperparameter tuning.}  
We denote the model trained on all \tofu{} authors as \textit{Target},  
and the model trained on $\D \setminus \Dforget$ as \textit{Ideal}.

Table~\ref{tab:tofu_forget10_OLMo-2-7B} presents the results on the OLMo-2-7B model.
As shown there,
$\method_{3691\text{B}}$, $\method_{3859\text{B}}$, and $\method{}_\text{last}$ achieve competitive results across all metrics.  
In fact, while each baseline typically fails on at least one metric, these \method{} variants remain competitive across all of them.  
For example, although RMU performs strongly overall, it shows low performance on \accrestor{},  
a metric that evaluates how well data-level unlearning is achieved.  
Similarly, while NPO attains reasonable performance, \method{} surpasses it for checkpoints that are within a hundred billion tokens of the unlearning target.  We also conduct the same experiments with the Llama-3.1-8B-Instruct model,  
with results shown in Table~\ref{tab:tofu_forget10_Llama-3.1-8B}.  
We observe that here too, \method{} variants obtain competitive results across all metrics,  
whereas other baselines often fail on at least one metric or underperform compared to \method{}.

\input{tables/msa_forget10_llama_8B}
\input{tables/msa_restor}


\paragraph{\method better recovers knowledge about real-world entities in \restor}

We evaluate \method{} on the \restor{} benchmark.  
A model is trained on \restor{} dataset, which introduces misinformation about a set of target entities,  
causing the model to lose its original knowledge and capabilities regarding those entities. Table~\ref{tab:restor_msa}
reports the results across both OLMo-2-7B models and Llama-3.1-8B-Instruct.  

For Llama-3.1-8B-Instruct, the ideal model, i.e.,
the model not trained on the \restor{} dataset,
achieves an accuracy of $64.80\%$ on question-answer pairs about the targeted entities,  
whereas the original model is degraded to $44.31\%$.  
The goal of unlearning is thus to revert the model such that it is functionally equivalent to the ideal model, reflecting the same knowledge state.  
As shown, while NPO and SatImp provide only limited recovery, \method{} achieves substantially better performance, recovering accuracy to a much greater extent. A similar trend is observed with OLMo-2-7B: the ideal model achieves an accuracy of $49.76\%$,
while the model continually trained on the \restor{} dataset drops to $37.60\%$.  
Here, SatImp yields only modest improvements, whereas \method{} variants provide strong recovery.    
We refer to Appendix~\ref{app:restor} for further experimental details.

\paragraph{\method is robust across diverse unlearning evaluation criteria from \muse-Books}

We evaluate unlearning algorithms on the \muse-Books benchmark, which considers diverse evaluation criteria for data-level unlearning, such as examining whether the unlearned model is susceptible to membership inference attacks featuring the unlearning target, which would indicate that the model still encodes information about the target (see a full description of \muse evaluation criteria in \S \ref{subsec:eval}).   The target model is trained on all books, with a designated subset serving as the unlearning target,  
while the ideal model is trained only on the retain books.  

Table~\ref{tab:muse_olmo2} reports results for the OLMo-2-7B model.  
As shown, \method{} performs strongly overall.  
Although $\method_{500\text{B}}$ and $\method_{2207\text{B}}$ show degraded performance in Knowledge Memorization on the retain set,  
\method{} variants leveraging closer checkpoints---$\method_{3691\text{B}}$, $\method_{3859\text{B}}$, and $\method_\text{last}$---achieve competitive results across all metrics.  
Notably, when evaluated with $\textsc{Min-K}\%$ and $\textsc{Min-K}\%^{++}$,  
two recent robust metrics for membership inference attacks,  
\method{} variants remain competitive and outperform other methods.  
This indicates stronger data-level unlearning, as unlearning documents are no longer identified as part of the training set.  
While RMU attains competitive performance, it is generally outperformed by \method{} variants.  Additional details on this experiment, as well as results on Llama models, are provided in Appendix~\ref{app:muse}. 

\input{tables/msa_muse_olmo2_7B}

\paragraph{\method can be effective even with infrequent checkpointing (within limits)}

We ask the question: how close in training does a checkpoint need to be to the unlearning target for \method to be effective, i.e., would the performance of \method suffer if a practitioner infrequently stores checkpoints?
For \restor{}, even early checkpoints---such as those trained on $500$B and $2207$B tokens---achieve competitive performance.  
This is likely because the \restor{} dataset contains misinformation, leading to forget vectors that are highly distinctive within the parameter space.  
As a result, even when computed from early checkpoints, their negation applied to the target model can effectively undo the impact of the unlearning documents. However, for \tofu{}, when \method leverages earlier checkpoints ($\method_{500\text{B}}$ and $\method{}_{2207\text{B}}$),  
the performance drops and competitive results cannot be maintained across all metrics. However, ($\method_{3691\text{B}}$ and $\method{}_{3859\text{B}}$) achieve competitive performance to the final chckpoint.
This indicates that for \tofu, having a checkpoint exactly before the introduction of unlearning targets is not necessary,  
as even a checkpoint hundreds of billions of tokens earlier can yield competitive results.  
However, \method{} with checkpoints too far away may lead to degraded unlearning performance. 

\paragraph{Unlearning as a tradeoff between objectives} We find that no single unlearning method proposed thus far clearly outperforms others on all metrics. 
For example, we find that \method{} aligns with the behavior of the ideal model.
In contrast, RMU performs well on \tofu, achieving higher \accforget{} and \accretain{}, but at the cost of very low \accrestor{}, as it often refuses to answer questions about authors in the forget set--- indeed such refusal \emph{could in itself be indicative of membership in a forget set}.
On the \muse benchmark, RMU achieves strong results (over-unlearning) on metrics such as exact and verbatim memorization, but falls behind \method{} on Privacy Leakage and $\textsc{Min-K\%}$.
Thus, \emph{practitioners must choose which unlearning method is applicable based on their priorities}: stronger data-level unlearning versus more aggressive removal of specific content without faithfully mimicking the ideal model. We argue that \method{} better supports a balance of several objectives for data-level unlearning, though it may not always be the most appropriate choice for other goals.

\paragraph{Unlearning when targets are introduced many tokens before the final checkpoint}
Recent work~\citep{yu2025impossibility} studies how the position of unlearning targets along the training trajectory affects unlearning,
and finds that introducing targets late in training is the most challenging regime.
This aligns with standard benchmarks and motivates our main evaluation setting.
Nevertheless, it is also important to study cases where the unlearning targets appear \emph{many tokens before} the final checkpoint $\theta_{\mathcal{D}}$.
To this end, we finetune Llama-3.2-1B-Instruct on \tofu and then continue finetuning on $\sim$20M tokens of C4, so the unlearning targets are not the last in training;
the ideal reference model is trained on the retain subset of \tofu and then finetuned on C4.
Table~\ref{tab:tofu_forget10_Llama-3.2-1B-msa_tofu_c4} reports the results and shows that \method{} remains effective: variants using checkpoints from before exposure to the forget set ($\method_{\text{base}}$ and $\method_{\text{instruct}}$) stay close to the ideal model.
In contrast, $\method_{\text{TOFU}}$ which uses the checkpoint \emph{after} finetuning on \tofu but \emph{before} the additional C4 finetuning--- underperforms on multiple metrics.
We refer to Appendix~\ref{app:tofu_c4} for more details.

\input{tables/msa_forget10_llama_1B_tofu_c4}

We include two additional investigations in the Appendix that probe settings beyond the standard benchmark setup.
Appendix~\ref{app:tofu_c4_tofu} studies unlearning under \emph{repeated exposure} to the forget data by training a model on \tofu{} + C4 + \tofu{}, where the targets appear multiple times.
We examines how checkpoint choice affects \method{} in this regime. In
Appendix~\ref{app:msa_baselines} we investigate whether existing unlearning baselines can similarly leverage intermediate checkpoints by extracting an update direction from an earlier checkpoint and applying it to the final target model, enabling a direct comparison between \method{} and checkpoint-augmented variants of prior methods.


%% file: tables/msa_forget10_llama_8B.tex
\begin{table}[t]
  \centering
  \footnotesize
    \caption{
    Comparison of unlearning algorithms on the \texttt{forget10} task from \tofu{}.  
    The target model is the Llama-3.1-8B-Instruct finetuned on all \tofu{} authors.  
    We report \dgap[1]{+100\%} when performance matches or exceeds that of the ideal model.  
    Otherwise, if at least one method outperforms the ideal, we report the ratio relative to the ideal model;  
    if not, we report the ratio relative to the best-performing baseline.  
    In these cases, values are shown as \dgap[0.8]{$X$\%}, where $X$ denotes the corresponding ratio.  
    \method{} variants achieve strong results, delivering superior or competitive performance across all metrics.  
    }
  \label{tab:tofu_forget10_Llama-3.1-8B}
  \begin{adjustbox}{width=\textwidth,center}
    \begin{tabular}{l|cc|cc|cc|cc|cc|cc}
      \toprule
      \multirow{2}{*}{Model} & \multicolumn{6}{|c|}{GPT-4o Judge Metrics $\uparrow$} & \multicolumn{6}{|c}{\tofu Metrics} \\
      \cmidrule(lr){2-7} \cmidrule(lr){8-13}
       & \multicolumn{2}{|c|}{\accforget} & \multicolumn{2}{|c|}{\accrestor} & \multicolumn{2}{|c|}{\accretain} & \multicolumn{2}{|c|}{Ext. Strength $\downarrow$} & \multicolumn{2}{|c|}{Model Utility $\uparrow$} & \multicolumn{2}{|c|}{ROUGE-L$_{\text{f}}$ $\downarrow$} \\
      \midrule \midrule
      Target & 0.03 &  & 0.02 &  & 1.00 &  & 0.98 &  & 0.57 &  & 0.99 &    \\
      Ideal & 0.98 &  & 0.98 &  & 1.00 &  & 0.07 &  & 0.60 &  & 0.39 &    \\
      \cmidrule(lr){1-13} 
      $\method_{\text{base}}$ & 0.82 & \dgap[0.951]{95.1\%} & 0.45 & \dgap[0.978]{97.8\%} & 0.92 & \dgap[0.922]{92.2\%} & 0.07 & \dgap[0.891]{89.1\%} & 0.78 & \dgap[1]{+100\%} & 0.40 & \dgap[0.995]{99.5\%} \\
      $\method_{\text{instruct}}$ & 0.82 & \dgap[0.956]{95.6\%} & 0.46 & \dgap[1.000]{100.0\%} & 0.91 & \dgap[0.917]{91.7\%} & 0.07 & \dgap[0.978]{97.8\%} & 0.57 & \dgap[0.949]{94.9\%} & 0.38 & \dgap[1]{+100\%} \\
      \cmidrule(lr){1-13}
      NPO & 0.75 & \dgap[0.872]{87.2\%} & 0.38 & \dgap[0.822]{82.2\%} & 0.83 & \dgap[0.834]{83.4\%} & 0.08 & \dgap[0.810]{81.0\%} & 0.58 & \dgap[0.956]{95.6\%} & 0.36 & \dgap[1]{+100\%}  \\
      RMU & 0.86 & \dgap[1.000]{100.0\%} & 0.12 & \dgap[0.254]{25.4\%} & 0.99 & \dgap[1.000]{100.0\%} & 0.07 & \dgap[0.868]{86.8\%} & 0.59 & \dgap[0.977]{97.7\%} & 0.19 & \dgap[1]{+100\%}  \\
      GradDiff & 0.49 & \dgap[0.573]{57.3\%} & 0.26 & \dgap[0.557]{55.7\%} & 0.88 & \dgap[0.879]{87.9\%} & 0.21 & \dgap[0.309]{30.9\%} & 0.64 & \dgap[1]{+100\%} & 0.45 & \dgap[0.872]{87.2\%}  \\
      Task Vector & 0.80 & \dgap[0.933]{93.3\%} & 0.27 & \dgap[0.578]{57.8\%} & 0.51 & \dgap[0.515]{51.5\%} & 0.03 & \dgap[1]{+100\%} & 0.53 & \dgap[0.887]{88.7\%} & 0.29 & \dgap[1]{+100\%} \\
      SatImp & 0.52 & \dgap[0.608]{60.8\%} & 0.28 & \dgap[0.616]{61.6\%} & 0.89 & \dgap[0.897]{89.7\%} & 0.15 & \dgap[0.445]{44.5\%} & 0.63 & \dgap[1]{+100\%} & 0.44 & \dgap[0.901]{90.1\%} \\
      UNDIAL & 0.46 & \dgap[0.538]{53.8\%} & 0.29 & \dgap[0.622]{62.2\%} & 0.84 & \dgap[0.847]{84.7\%} & 0.08 & \dgap[0.797]{79.7\%} & 0.65 & \dgap[1]{+100\%} & 0.41 & \dgap[0.951]{95.1\%} \\
      \bottomrule
    \end{tabular}
  \end{adjustbox}
\end{table}

%% file: tables/msa_restor.tex


\begin{table}[b]
\centering
\caption{
Performance of unlearning algorithms on \restor{} benchmark, measured by accuracy on $1051$ question–answer pairs of \restor{} across both Llama-3.1-8B-Instruct and OLMo-2-7B models.
}
\label{tab:restor_msa}
\begin{adjustbox}{width=\textwidth}
\begin{tabular}{l|cc|ccccc|ccccc}
\toprule
{Model} & {Target} & {Ideal} & \multicolumn{5}{c|}{{\method}} & {NPO} & {GradDiff} & {Task Vector} & {SatImp} & {RMU} \\
\midrule \midrule
\multirow{2}{*}{Llama-3.1-8B}
& \multirow{2}{*}{44.31}
& \multirow{2}{*}{64.80}
& \multicolumn{2}{c}{\(\method_{\text{base}}\)}
& \multicolumn{3}{c|}{\(\method_{\text{instruct}}\)}
& \multirow{2}{*}{48.45}
& \multirow{2}{*}{26.08}
& \multirow{2}{*}{44.50}
& \multirow{2}{*}{49.19}
& \multirow{2}{*}{41.47}
\\
& & & \multicolumn{2}{c}{59.40} & \multicolumn{3}{c|}{\textbf{63.95}} & & & & & \\
\midrule
\multirow{2}{*}{OLMo-2-7B}
& \multirow{2}{*}{37.60}
& \multirow{2}{*}{49.76}
& \(\method_{500\text{B}}\)
& \(\method_{2207\text{B}}\)
& \(\method_{3691\text{B}}\)
& \(\method_{3859\text{B}}\)
& \(\method_{\text{last}}\)
& \multirow{2}{*}{34.73}
& \multirow{2}{*}{21.28}
& \multirow{2}{*}{38.47}
& \multirow{2}{*}{40.25}
& \multirow{2}{*}{36.00}
\\
& & & 45.67 & 46.21 & 47.27 & 47.64 & \textbf{47.80} & & & & & \\
\bottomrule
\end{tabular}
\end{adjustbox}
\end{table}

%% file: tables/msa_muse_olmo2_7B.tex
\begin{table}[t]
  \centering
  \footnotesize
  \caption{
Comparison of unlearning algorithms on the \muse-Books benchmark.  
The target model is OLMo-2-7B finetuned on all \muse books.
We report \dgap[1]{+100\%} when performance matches or exceeds that of the ideal model.  
Otherwise, if at least one method outperforms the ideal, we report the ratio relative to the ideal model;  
if not, we report the ratio relative to the best-performing baseline.  
In these cases, values are shown as \dgap[0.8]{$X$\%}, where $X$ denotes the corresponding ratio.   
}
  \label{tab:muse_olmo2}

  \begin{adjustbox}{width=\textwidth,center}
    \begin{tabular}{l|cc|cc|cc|cc|cc|cc|cc}
      \toprule
      Model & \multicolumn{2}{|c|}{Ext. Strength $\downarrow$} & \multicolumn{2}{|c|}{Exact Mem $\downarrow$} & \multicolumn{2}{|c|}{VerbMem $\Dforget$ $\downarrow$} & \multicolumn{2}{|c|}{$\textsc{Min-K}\%$ $\downarrow$} & \multicolumn{2}{|c|}{$\textsc{Min-K}\%^{++}$ $\downarrow$} & \multicolumn{2}{|c|}{KnowMem $\Dretain$ $\uparrow$} & \multicolumn{2}{|c}{PrivLeak $\rightarrow 0$} \\
      \midrule \midrule
      Target & 0.43 &  & 0.94 &  & 0.49 &  & 1.00 &  & 1.00 &  & 0.62 &  & \multicolumn{2}{|c}{{\small -100.00}} \\
      Ideal & 0.02 &  & 0.54 &  & 0.17 &  & 0.45 &  & 0.39 &  & 0.67 &  & \multicolumn{2}{|c}{{\small 0.00}} \\
      \cmidrule(lr){1-15}
      $\method_{500\text{B}}$ & 0.01 & \dgap[1]{+100\%} & 0.41 & \dgap[1]{+100\%} & 0.12 & \dgap[1]{+100\%} & 0.14 & \dgap[1]{+100\%} & 0.09 & \dgap[1]{+100\%} & 0.51 & \dgap[0.774]{77.4\%} & \multicolumn{2}{|c}{{\small 56.38}} \\
      $\method_{2207\text{B}}$ & 0.01 & \dgap[1]{+100\%} & 0.37 & \dgap[1]{+100\%} & 0.10 & \dgap[1]{+100\%} & 0.04 & \dgap[1]{+100\%} & 0.01 & \dgap[1]{+100\%} & 0.45 & \dgap[0.691]{69.1\%} & \multicolumn{2}{|c}{{\small 74.05}} \\
      $\method_{3691\text{B}}$ & 0.02 & \dgap[1]{+100\%} & 0.51 & \dgap[1]{+100\%} & 0.15 & \dgap[1]{+100\%} & 0.30 & \dgap[1]{+100\%} & 0.21 & \dgap[1]{+100\%} & 0.63 & \dgap[0.955]{95.5\%} & \multicolumn{2}{|c}{{\small 27.63}} \\
      $\method_{3859\text{B}}$ & 0.02 & \dgap[1]{+100\%} & 0.51 & \dgap[1]{+100\%} & 0.15 & \dgap[1]{+100\%} & 0.23 & \dgap[1]{+100\%} & 0.16 & \dgap[1]{+100\%} & 0.59 & \dgap[0.905]{90.5\%} & \multicolumn{2}{|c}{{\small \textbf{23.45}}} \\
      $\method_{\text{last}}$ & 0.02 & \dgap[0.998]{99.8\%} & 0.55 & \dgap[0.970]{97.0\%} & 0.16 & \dgap[1]{+100\%} & 0.37 & \dgap[1]{+100\%} & 0.22 & \dgap[1]{+100\%} & 0.65 & \dgap[1.000]{100.0\%} & \multicolumn{2}{|c}{{\small \textbf{14.67}}} \\
      \cmidrule(lr){1-15}
      NPO & 0.02 & \dgap[0.881]{88.1\%} & 0.64 & \dgap[0.840]{84.0\%} & 0.15 & \dgap[1]{+100\%} & 1.00 & \dgap[0.448]{44.8\%} & 0.99 & \dgap[0.392]{39.2\%} & 0.62 & \dgap[0.950]{95.0\%} & \multicolumn{2}{|c}{{\small -99.93}} \\
      RMU & 0.01 & \dgap[1]{+100\%} & 0.06 & \dgap[1]{+100\%} & 0.08 & \dgap[1]{+100\%} & 0.55 & \dgap[0.820]{82.0\%} & 0.47 & \dgap[0.833]{83.3\%} & 0.64 & \dgap[0.977]{97.7\%} & \multicolumn{2}{|c}{{\small \textbf{-17.83}}} \\
      GradDiff & 0.01 & \dgap[1]{+100\%} & 0.20 & \dgap[1]{+100\%} & 0.01 & \dgap[1]{+100\%} & 0.50 & \dgap[0.895]{89.5\%} & 0.45 & \dgap[0.870]{87.0\%} & 0.45 & \dgap[0.689]{68.9\%} & \multicolumn{2}{|c}{{\small \underline{-9.47}}} \\
      Task-Vector & 0.01 & \dgap[1]{+100\%} & 0.46 & \dgap[1]{+100\%} & 0.13 & \dgap[1]{+100\%} & 0.92 & \dgap[0.489]{48.9\%} & 0.95 & \dgap[0.408]{40.8\%} & 0.48 & \dgap[0.735]{73.5\%} & \multicolumn{2}{|c}{{\small -84.30}} \\
      SatImp & 0.37 & \dgap[0.049]{4.9\%} & 0.93 & \dgap[0.576]{57.6\%} & 0.43 & \dgap[0.401]{40.1\%} & 1.00 & \dgap[0.448]{44.8\%} & 1.00 & \dgap[0.388]{38.8\%} & 0.62 & \dgap[0.947]{94.7\%} & \multicolumn{2}{|c}{{\small -100.00}} \\
      UNDIAL & 0.02 & \dgap[0.785]{78.5\%} & 0.64 & \dgap[0.836]{83.6\%} & 0.16 & \dgap[1]{+100\%} & 1.00 & \dgap[0.448]{44.8\%} & 1.00 & \dgap[0.388]{38.8\%} & 0.53 & \dgap[0.804]{80.4\%} & \multicolumn{2}{|c}{{\small -100.00}} \\
      \bottomrule
    \end{tabular}
  \end{adjustbox}
\end{table}

%% file: tables/msa_forget10_llama_1B_tofu_c4.tex
\begin{table}[t]
  \centering
\caption{Comparison of MSA variants on \tofu{} (\texttt{forget10}).
  In this scenario, unlearning targets are not introduced at the very end of the training pipeline; instead, the model later undergoes finetuning on a subset of C4 for 2 epochs.
  \method{} variants that use checkpoints prior to the unlearning targets, i.e., $\method{}_{\text{base}}$ and $\method{}_{\text{instruct}}$, show acceptable performance, achieving values near the ideal model.}
  \label{tab:tofu_forget10_Llama-3.2-1B-msa_tofu_c4}
  \setlength{\tabcolsep}{4pt}
  \begin{adjustbox}{width=\textwidth,center}
    \begin{tabular}{l|cc|cc|cc|cc|cc|cc|cc}
      \toprule
      \multirow{2}{*}{Model} & \multicolumn{6}{|c|}{GPT-4o Judge Metrics $\uparrow$} & \multicolumn{8}{|c}{\tofu Metrics} \\
      \cmidrule(lr){2-7} \cmidrule(lr){8-15}
       & \multicolumn{2}{|c|}{\accforget} & \multicolumn{2}{|c|}{\accrestor} & \multicolumn{2}{|c|}{\accretain} & \multicolumn{2}{|c|}{ES on $\Dforget$ $\downarrow$} & \multicolumn{2}{|c|}{Model Utility $\uparrow$} & \multicolumn{2}{|c|}{ROUGE-L$_{\text{f}}$ $\downarrow$} & \multicolumn{2}{|c}{Forget Quality $\uparrow$} \\
      \midrule \midrule
      Target & 0.48 &  & 0.24 &  & 0.66 &  & 0.19 &  & 0.55 &  & 0.49 &  & \multicolumn{2}{|c}{9.34e-13} \\
      Ideal & 0.83 &  & 0.98 &  & 0.69 &  & 0.07 &  & 0.55 &  & 0.38 &  & \multicolumn{2}{|c}{1} \\
      \cmidrule(lr){1-15}
      $\method_{\text{base}}$ & 0.79 & \dgap[0.955]{95.5\%} & 0.39 & \dgap[0.876]{87.6\%} & 0.68 & \dgap[0.982]{98.2\%} & 0.06 & \dgap[1]{+100\%} & 0.53 & \dgap[0.978]{97.8\%} & 0.34 & \dgap[1]{+100\%} & \multicolumn{2}{|c}{0.42} \\
      $\method_{\text{instruct}}$ & 0.83 & \dgap[1.000]{100.0\%} & 0.45 & \dgap[1]{+100\%} & 0.70 & \dgap[1]{+100\%} & 0.06 & \dgap[1]{+100\%} & 0.55 & \dgap[1]{+100\%} & 0.36 & \dgap[1]{+100\%} & \multicolumn{2}{|c}{0.70} \\
      $\method_{{\tofu}}$ & 0.73 & \dgap[0.882]{88.2\%} & 0.37 & \dgap[0.826]{82.6\%} & 0.70 & \dgap[1]{+100\%} & 0.08 & \dgap[0.802]{80.2\%} & 0.57 & \dgap[1]{+100\%} & 0.33 & \dgap[1]{+100\%} & \multicolumn{2}{|c}{\rej{1.10e-09}} \\
      \bottomrule
    \end{tabular}
  \end{adjustbox}
\end{table}

%% file: sections/conclusion.tex
\section{Conclusion}

We introduce \method{}, a new method for machine unlearning that leverages intermediate model checkpoints to estimate and undo the influence of undesirable data.
By casting unlearning as arithmetic in parameter space, \method{} enables targeted forgetting.
Across \tofu, \muse-Books and \restor benchmarks,
\method outperforms prior methods over a variety of metrics, 
achieving superior forgetting, recovery, and utility preservation—even when unlearning directions are computed from early checkpoints.
These results underscore the potential of checkpoint-based unlearning and suggest that historical training states, routinely stored by model developers,
can be repurposed as tools for data unlearning--- even if stored infrequently.
We hope \method inspires further research into lightweight, generalizable, and interpretable unlearning techniques for large language models.


\section*{Acknowledgement}
This project was supported in part by a grant from an NSF CAREER AWARD 1942230, the ONR PECASE grant N00014-25-1-2378, ARO’s Early Career Program Award 310902-00001, Army Grant No. W911NF2120076, the NSF award CCF2212458, NSF Award No. 2229885 (NSF Institute for Trustworthy AI in Law and Society, TRAILS), a MURI grant 14262683, DARPA AIQ grant HR00112590066 and an award from meta 314593-00001.

\section*{Ethics Statement}
We adhere to the ICLR Code of Ethics and design this work to support responsible data governance by enabling post-hoc removal of targeted training data.
Our method, Model State Arithmetic (MSA), computes a “forget vector” from a prior checkpoint and applies it to the trained model to reduce the influence of specified data while preserving overall capability (\S\ref{sec:methodology}).
We motivate unlearning in the context of privacy, copyright, and regulatory deletion requests, and discuss practical guardrails for safe use (\S\ref{sec:intro}).
All experiments use public unlearning benchmarks—\tofu, \restor, and \muse-Books—following their established protocols; no new human-subject data were collected (\S\ref{sec:experimiments}), \cite{maini2024tofu, rezaei2024restor, shi2024muse}.
We acknowledge potential risks (e.g., erasing beneficial safety behaviors) and mitigate it by coupling forgetting with retention objectives and by reporting utility beyond the forget set (\S\ref{sec:experimiments}).

\section*{Reproducibility Statement}
We provide the algorithmic specification of MSA, including the update rule
\(\theta_{\text{unlearn}}=\theta_\D-\alpha\,\vec{\theta}_f\,(+\,\beta\,\vec{\theta}_r)\),
with implementation details and checkpoint usage (\S\ref{sec:methodology}). Datasets, splits, prompts, and evaluation protocols for \tofu, \restor, and \muse-Books are described in the main text (\S\ref{sec:experimiments}) and the Appendix.
Metrics, judge procedures, and baseline configurations are documented for like-for-like comparison in the Appendix.
Code is also available in Github.


%% file: appendix.tex
\newpage

\section{Extended Related Work}
\label{app:related_work}

\paragraph{Amnesiac Machine Unlearning \cite{graves2021amnesiac}.}
Although conceptually related to our approach, since it also exploits information from the model's training trajectory, amnesiac machine unlearning faces two key limitations that make it impractical for large language models:

First, it requires logging and storing the full parameter update vector for every training step whose batch might later be subject to deletion, along with a record of which examples appear in which batches. In realistic deletion scenarios, this implies maintaining an \(O(\#\text{steps} \times |\theta|)\) log of updates, which is vastly larger than the handful of checkpoints typically retained in LLM training and becomes prohibitive at the scales at which large language models are trained (multi-billion-parameter models trained on trillions of tokens). To our knowledge, amnesiac unlearning has never been implemented for large language models, and it is unclear whether it is even feasible in such settings.

Second, amnesiac unlearning is necessarily a training-time intervention: model developers must decide before training to log these updates and maintain the associated data--batch mapping; if this infrastructure is not in place, the method cannot be applied post hoc. By contrast, \method{} requires only access to intermediate checkpoints that are already routinely saved in standard LLM training pipelines. Combined, these considerations make \method{} more practical for large language models and enable post-hoc unlearning, as demonstrated by our application to existing models such as OLMo, without any prior modifications or special preparation during training.

\paragraph{Unrolling SGD \cite{thudi2022unrolling}.}
The Unrolling SGD framework studies approximate machine unlearning by analyzing SGD and proposing \emph{verification error}, defined as the distance in weight space between an approximately unlearned model and the ideal retrained model. The authors introduce (i) single-gradient unlearning, which uses the model checkpoint before training on the forget example together with a single gradient step to approximate removal, and (ii) a training-time regularizer that constrains the SGD trajectory to make future unlearning requests easier. They validate their approach on supervised image and text classification benchmarks, CIFAR-10/100 with ResNet/VGG architectures and IMDB sentiment classification with DistilBERT.

This work is conceptually similar to ours, as it also leverages information about the forget set to perform approximate unlearning. However, our approach differs in several important respects. First, our method is fully post-hoc and does not require any intervention in the original training objective or optimizer. Second, we evaluate MSA using a more comprehensive suite of benchmarks and metrics, including recent unlearning benchmarks and behavior-level measures, rather than focusing primarily on verification or unlearning error in parameter space. Third, we apply \method{} at LLM scale, with large models trained on billions of tokens. In contrast to the experimental setup of~\citep{thudi2022unrolling}, which assumes access to a model checkpoint taken immediately before the introduction of the unlearning targets, we conduct real-scale experiments using checkpoints that may lie billions of tokens before the forget set. Finally, the empirical performance reported in~\citep{thudi2022unrolling} appears to degrade when the training-time regularization term is removed, whereas our method achieves strong empirical performance in a purely post-hoc setting without any modification to the original training process.

It is worth noting that we are not the first to look at using a previous model state to compute gradients for forgetting, and \citep{thudi2022unrolling} uses vectors derived from a pretrained model state (similar to $\method{}_\text{base}$), and an initial model state.

\paragraph{Rewind-to-Delete \cite{mu2024rewind}.}
Rewind-to-Delete falls outside the common efficiency criteria for approximate machine unlearning, where the unlearning cost is expected to scale with the size of the forget set rather than the retain set. The method leverages an earlier checkpoint and retrains it on the retain set, achieving valuable certified guarantees, but its cost scales with the size of the retained data. Consequently, it does not fit within the typical efficiency regime of approximate unlearning methods whose complexity is \(O(|\mathcal{D}_\text{f}|)\), such as \method{}, NPO, and GA in the LLM setting.


On overall, we propose \method{} as an efficient approximate unlearning algorithm whose runtime scales as $O(|\mathcal{D}_\text{f}|)$,
similar to other efficient approximate unlearning methods, while explicitly leveraging model checkpoints under the constraints of LLM training pipelines. Unlike prior approaches~\cite{thudi2022unrolling, graves2021amnesiac}, \method{} does not require storing training parameter updates or having control over the training objective or optimizer, and instead operates purely post hoc on existing checkpoints. Across multiple benchmarks and evaluation metrics, MSA achieves competitive, and often superior, performance compared to prior baselines. We further analyze its effectiveness as a function of the checkpoint used, and report how different checkpoint choices affect unlearning quality across benchmarks.

\section{GPT-4o for \tofu Metrics}
\label{app:judge}

We utilize GPT-4o to evaluate model outputs by identifying which candidate response 
is most semantically similar to the model's output. 
Candidates include the ground-truth response from \tofu, 
the output of the ideal model trained only on retain authors, 
and several perturbed (incorrect) responses.
We use the following prompt to instruct the judge model:

\begin{lstlisting}{}
@System Prompt:@

You are a factual consistency evaluator. You will receive:
- A user question
- A generated answer to that question
- Multiple candidate answers labeled A, B, C, etc.
Task:\n
Before anything else, check the generated answer:
- If it is incoherent, nonsensical, gibberish, or fails to convey any real facts,
    immediately reply with Z.
Otherwise, proceed:
- Select exactly one letter (A, B, C, ...) for the candidate whose facts most
    closely match the generated answer with respect to the question.
- Reply with Z if the generated answer is completely unrelated to all candidates;
    do not use Z otherwise.
- If two or more candidates tie for highest factual similarity, choose the one
    with the earliest letter (A before B, B before C, etc.).
Always reply with exactly one letter (A, B, C, ... or Z) and no additional text.

\end{lstlisting}

\begin{lstlisting}{}
@User Prompt:@

Question:
{[input text]}

Generated answer:
{[generated text]}

Candidates:
{[random_shuffle(ground truth, ideal model output, *perturbed answers)]}

Which candidate (A, B, C, ...) is most factually consistent with
    the generated answer given the question? 
Reply with the single letter only.

\end{lstlisting}

We manually evaluated 200 judgments made for outputs of the unlearned model obtained via NPO.  
The GPT-4o-based judge \textbf{achieved an accuracy of $96\%$}—that is, in $96\%$ of cases,  
the option selected as most similar matched the choice a human evaluator would have made.  
Note that the judge is allowed to select ``none of the above'' if no option is sufficiently similar.  
Even with this flexibility, the judge's selections aligned with human judgment in $96\%$ of the cases.

\input{figures/metrics_retain}

\subsection{Limitations of ROUGE-L for Forgetting Evaluation}
\label{app:rouge}

In Figure~\ref{fig:metric_samples} and Figure~\ref{fig:metric_samples_retain}, we provide qualitative examples to illustrate a key limitation of using ROUGE-L (or other metrics considering all tokens of ground-truth and output) for evaluating machine unlearning.
Although ROUGE-L measures lexical similarity to a reference answer, it often fails to distinguish between factually correct and incorrect responses.
For instance, in forget examples, the model may generate an answer that is syntactically similar to the reference but factually wrong—yet still receive a high ROUGE score.
Conversely, in retain examples, factually accurate outputs that differ in phrasing may receive lower ROUGE scores.

\section{Experiments on \tofu}
\label{app:tofu}

\input{tables/msa_forget01_llama_1B}
\input{tables/msa_forget05_llama_1B}
\input{tables/msa_forget10_llama_1B}
\input{tables/msa_forget10_llama_8B_appendix}

In this section, we provide additional experimental details  
for running the \tofu{} experiments.  
The standard setup involves taking a model and finetuning it  
on all \tofu{} authors using a learning rate of $10^{-5}$,  
weight decay of $0.01$, one warm-up epoch, and a total of $5$ training epochs.  
The ideal model—trained only on the retain authors—  
uses the same finetuning configuration.  
All experiments are run on $2$ A100 GPUs.  

We use Llama-3.1-8B-Instruct, Llama-3.2-1B-Instruct,  
and the final checkpoint of stage 1 pretraining of OLMo-2-7B as the base models for training on \tofu{}.  

\subsection{Forget Quality}
We note that although Forget Quality was introduced by \citet{maini2024tofu},  
we found the metric to be highly sensitive, often producing very low values that can hinder clear comparison in the main tables.  
Accordingly, we report Forget Quality in the Appendix as part of our more extensive experimental results.

\subsection{Obtaining Forget and Retain Vectors}

We finetune the checkpoint $C$ prior to the exposure to the \tofu{} dataset for $5$ epochs to obtain the forget vector.  
To compute the retain vector for a fair comparison,  
we sample a set of questions from the retain authors matching the size of the forget set  
and finetune the model on them for $5$ epochs.  

\subsection{Choosing Hyperparameters of \method{} and Baselines}

We split our evaluation dataset into validation ($15\%$) and test ($85\%$) sets.  
To find the best set of hyperparameters in \tofu{} experiments,  
we define a validation score as the geometric mean of several metrics on the validation set:

$$
\text{Score} = e^{\frac{(\text{Model Utility})^2 (\text{Acc}_{\text{forget}}) (\text{Acc}_{\text{recover}})^2 (\text{Acc}_{\text{retain}}) (1 - \text{extraction strength})^2}{8}}
$$

This score ensures that the chosen hyperparameters balance a good trade-off across metrics,  
with greater emphasis on \accrestor (as it measures ideal data-level unlearning),  
Model Utility (to ensure the model remains useful on related tasks),  
and extraction strength (a robust metric for unlearning evaluation).  

\paragraph{\texttt{forget10} -- Llama-3.1-8B-Instruct}  

For \method{} and Task Vector, $\alpha \in \{0.5, 0.75, 1.0, 1.25, 1.5, 3.0\}$ and $\beta \in \{0.5, 1.0, 1.5\}$,  
yielding $15$ cases in total.  
The best-performing $\alpha$ and $\beta$ are selected for final evaluation.  

For the baselines, we perform unlearning for $5$ epochs and evaluate each checkpoint after every epoch:  

\begin{itemize}
    \item {NPO:} $\lambda \in \{2, 4\}$, learning rate $\in \{10^{-5}, 2 \times 10^{-5}\}$,  
    for $5 \times 2 \times 2 = 20$ settings.  
    \item {GradDiff:} $\lambda \in \{2, 4\}$, learning rate $10^{-5}$,  
    for $5 \times 2 = 10$ settings.  
    \item {UNDIAL:} $\lambda \in \{1, 2, 4\}$, learning rate $2 \times 10^{-5}$,  
    for $5 \times 3 = 15$ settings.  
    \item {SatImp:} $\gamma \in \{4, 8\}$, learning rate $10^{-5}$, $\beta_1 = 5$, $\beta_2 = 1$,  
    for $5 \times 2 = 10$ settings.  
    \item {RMU:} $\lambda \in \{2, 4\}$, learning rate $10^{-5}$,  
    for $5 \times 2 = 10$ settings.  
\end{itemize}

\paragraph{\texttt{forget01}, \texttt{forget05}, and \texttt{forget10} -- Llama-3.2-1B-Instruct}  

For the smaller Llama-3.2-1B-Instruct model, we can perform a more extensive hyperparameter search.  
For \method{} and Task Vector, we set $\alpha \in \{0.5, 0.75, 1.25, 1.5, 3.0\}$ and $\beta \in \{0.5, 0.75, 1.0, 1.25, 1.5\}$,  
yielding $25$ cases in total.  
The best-performing $\alpha$ and $\beta$ are used for the final evaluation.  

For baselines, we perform unlearning for $10$ epochs and evaluate each checkpoint after every epoch:  

\begin{itemize}
    \item {NPO:} $\lambda \in \{2, 4, 8\}$, learning rate $\in \{10^{-5}, 2 \times 10^{-5}\}$,  
    for $3 \times 2 \times 10 = 60$ settings.  
    \item {GradDiff:} $\lambda \in \{1, 2, 4\}$, learning rate $\in \{10^{-5}, 2 \times 10^{-5}\}$,  
    for $3 \times 2 \times 10 = 60$ settings.  
    \item {UNDIAL:} $\lambda \in \{1, 2, 4\}$, learning rate $\in \{10^{-5}, 2 \times 10^{-5}\}$,  
    for $3 \times 2 \times 10 = 60$ settings.  
    \item {SatImp:} $\gamma \in \{0.1, 1.0, 4.0\}$, learning rate $\in \{10^{-5}, 2 \times 10^{-5}\}$, $\beta_1 = 5$, $\beta_2 = 1$,  
    for $3 \times 2 \times 10 = 60$ settings.  
    \item {RMU:} $\alpha \in \{1, 2, 4\}$, learning rate $10^{-5}$,  
    for $3 \times 10 = 30$ settings.  
\end{itemize}

Results for Llama-3.2-1B-Instruct are reported in  
Table~\ref{tab:tofu_forget01_Llama-3.2-1B} for \texttt{forget01},  
Table~\ref{tab:tofu_forget05_Llama-3.2-1B} for \texttt{forget05}, and  
Table~\ref{tab:tofu_forget10_Llama-3.2-1B} for \texttt{forget10}.  

\section{Experiments on \restor}
\label{app:restor}

We follow the procedure described by~\citet{rezaei2024restor},  
starting with Llama-3.1-8B-Instruct and OLMo-2-7B,  
and finetune them on \restor{} for $5$ epochs using a learning rate of $10^{-5}$,  
weight decay of $0.01$, and $1$ warm-up epoch.  
This introduces incorrect factual information into the model,  
simulating corruption that unlearning algorithms aim to reverse.  
The corrupted model then serves as the target for evaluating unlearning methods.  

To tune hyperparameters, we hold out $10\%$ of the \restor{} questions as a validation set  
and evaluate accuracy on this subset.  
\method{} does not use any retain set in this setup,  
while other algorithms rely on C4 as their retain set to preserve model utility.  

We evaluate \method{} with $\alpha \in \{0.75, 1.0, 1.5, 2.0\}$.  
For baselines, we perform unlearning for $5$ epochs,  
evaluating the model on the validation set after each epoch.  
We set $\alpha=4$ and a learning rate of $10^{-5}$ for GradDiff, NPO, RMU, and UNDIAL,  
and $\gamma=4$, $\beta_1=5$, $\beta_2=1$ for SatImp.  

\section{Experiments on \muse-Books}
\label{app:muse}

We follow the procedure described in~\citet{shi2024muse},  
finetuning each model for $10$ epochs with a constant learning rate of $10^{-5}$.  
All experiments are run on $2$ A100 GPUs.  

We use the OLMo-2-7B checkpoint as before for finetuning on \muse{} books,  
as well as Llama-3-8B (we take a pretrained base model rather than instruct model to be consistent with \citet{shi2024muse})

\paragraph{Forget and Retain Vectors}  
To obtain forget and retain vectors for \method{},  
we use a checkpoint $C$ (depending on the model used).  
The forget vector is obtained by training on the unlearning target books for $5$ epochs  
with a learning rate of $10^{-5}$, weight decay of $0.01$, and $1$ warm-up epoch.  
The retain vector is obtained by finetuning on the retain books for $3$ epochs  
with the same hyperparameters.  
Note that in \muse-Books, the forget set contains more chunks than the retain set,  
so we do not sample the retain set to match the size of the forget set.  

\paragraph{Hyperparameter Selection}  
We split the \muse-Books benchmark into validation ($15\%$) and test ($85\%$) sets.  
As in the \tofu{} experiments, we design a validation score to balance trade-offs across metrics:  

\begin{align*}
    \text{Score} = e^{
    \frac{(1 - \textsc{Min-K}\%)
          (1 - \textsc{Min-K}\%^{++})
          (1 - \text{VerbMem}_{\text{f}})
          (1 - \text{KnowMem}_{\text{r}})^2
          (1 - \text{extraction strength})^2
          (1 - \text{exact memorization})}
         {8}}
\end{align*}

We place stronger emphasis on extraction strength  
and knowledge memorization of the retain set,  
to ensure that knowledge of the retain set is preserved in the unlearned model.  

\paragraph{Unlearning Algorithms}  
For \method{}, we set $\alpha \in \{0.75, 1.0, 1.5\}$ and $\beta \in \{0, 0.75, 1.0, 1.5\}$,  
selecting the configuration that maximizes the validation score for test evaluation.  

For baselines, we set $\lambda=4$ for NPO, GradDiff, RMU, and UNDIAL,  
and $\gamma=4$ for SatImp.  
We perform unlearning for $5$ epochs, evaluating each checkpoint on the validation set.  

Results for Llama-3.1-8B (as in~\citet{shi2024muse}) are shown in Table~\ref{tab:muse_llama2}.  

\input{tables/msa_muse_llama_7B}

We note that $\text{KnowMem}_{f}$, i.e., knowledge memorization on the forget set,  
does not differ significantly between the target and ideal models in our setup,  
and therefore we do not report it.  

\section{Unlearning Targets Introduced Many Tokens Before the Final Checkpoint}
\label{app:tofu_c4}

Most existing machine unlearning benchmarks \cite{maini2024tofu, rezaei2024restor, shi2024muse} typically assume that the unlearning targets are introduced at the end of training,
and we largely follow this setup to enable fair comparison with prior unlearning algorithms. Recent work~\citep{yu2025impossibility} studies how the position of the unlearning targets in the training trajectory affects unlearning performance, and shows that the most challenging setting is indeed when the targets are introduced late in training.
This aligns with the existing benchmarks and supports our choice to evaluate \method{} (and baselines) under this challenging regime.

Nevertheless, it is also important to understand scenarios in which the model is asked to forget information that was seen many tokens before the final checkpoint~$\theta_{\mathcal{D}}$.
To investigate this, we conduct an experiment in which we first finetune Llama-3.2-1B-Instruct on \tofu and then further finetune it on approximately $20$M tokens of C4.
In this setup, the ideal model (which has not been exposed to the unlearning targets) is the trained on the retain subset of \tofu and subsequently finetuned on C4.

Table~\ref{tab:tofu_forget10_Llama-3.2-1B-msa_tofu_c4} reports the results in this scenario. As seen there, \method{} variants that use checkpoints taken before the introduction of the unlearning targets, namely $\method{}_{\text{base}}$ and $\method{}_{\text{instruct}}$, remain effective and achieve values close to the ideal model, even though the unlearning targets now lie many tokens before the final checkpoint.
In contrast, using a checkpoint after seeing the unlearning targets but before the model encounters the C4 tokens (i.e., $\method{}_{\text{TOFU}}$) underperforms on multiple metrics.

These results provide empirical evidence that \method{} can still work well when the model is asked to forget information learned a significant number of tokens earlier, while reinforcing our earlier observation that checkpoints taken after exposure to the forget set are less suitable for constructing effective unlearning updates.

\section{Unlearning with Repeated Exposure to \tofu}
\label{app:tofu_c4_tofu}
\input{tables/msa_forget10_llama_1B_tofu_c4_tofu}

We next consider a setting where the forget data appears multiple times in the training corpus and is not always close to the final checkpoint~$\theta_{\mathcal{D}}$. To simulate this scenario, we start from Llama-3.2-1B-Instruct, first finetune it on \tofu, then train it on a subset of C4 (approximately 20M tokens), and finally finetune again on \tofu. This final model (\tofu{} + C4 + \tofu) is the target of unlearning. The ideal model in this setup is trained on \tofu{} retain, then C4, then \tofu{} retain again.

Table~\ref{tab:tofu_forget10_Llama-3.2-1B-msa_tofu_c4_tofu} reports the empirical results in this configuration. There are five natural checkpoints at which to apply \method{}: (1) the base model, (2) the instruct model, (3) the model after the first \tofu{} stage, (4) the model after \tofu{} + C4, and (5) the final model after \tofu{} + C4 + \tofu. As seen in the table, when \method{} leverages checkpoints that precede any exposure to \tofu{} (i.e., $\method{}_{\text{base}}$ and $\method{}_{\text{instruct}}$), it achieves strong performance, with values close to the ideal model. In contrast, using checkpoints that have already seen \tofu{} systematically underperforms.

This pattern suggests that, when the unlearning target is duplicated, the most effective checkpoints for \method{} are those prior to the first exposure of the model to the unlearning target.

\section{Augmenting Baselines with Intermediate Checkpoints}
\label{app:msa_baselines}

To investigate whether standard unlearning algorithms can also benefit from intermediate checkpoints, we apply these methods to earlier model states and then reuse the resulting update directions on the target model. More specifically, let $\theta_0$ be an intermediate checkpoint. We apply a baseline unlearning algorithm starting from $\theta_0$, obtaining a model $\theta_1$. We then extract the change direction $\theta_1 - \theta_0$ and apply it to the target model $\theta_{\mathcal{D}}$ with a tunable scalar $\alpha$, yielding
\begin{equation}
    \theta_{\text{unlearn}} = \theta_{\mathcal{D}} + \alpha (\theta_1 - \theta_0).
\end{equation}
We select the optimal value of $\alpha$ via validation search, as we do for other methods.

Table~\ref{tab:tofu_forget10_Llama-3.2-1B-msa-baselines} reports experimental results on the TOFU \texttt{forget10} task with Llama-3.2-1B, where unlearning algorithms are augmented with model checkpoints following the above procedure. For example, when applying NPO, we denote $\text{NPO}_{\text{base}}$ and $\text{NPO}_{\text{instruct}}$ for NPO applied to the pretrained base model and the instruct model, respectively, while \textsc{NPO} alone refers to the case where it is applied to the target model.

As seen in Table~\ref{tab:tofu_forget10_Llama-3.2-1B-msa-baselines}, these algorithms do not benefit from leveraging intermediate checkpoints in this way; they are outperformed by our method and typically exhibit degraded performance compared to their standard variants applied directly to the unlearning targets.

\input{tables/msa_forget10_llama_1B_msa_baselines}

\section{Potential Overlap with Pretraining Data}
\label{app:pretraining_overlap}

A potential limitation of our evaluation is that some of the datasets used may overlap with the pretraining data of the underlying models. In particular, if evaluation examples are present (or closely paraphrased) in the pretraining corpus, this could confound the interpretation of memorization and unlearning performance.

We note that \tofu{} and \restor{} are both synthetic datasets that are unlikely to be part of the pretraining data. In fact, \tofu{} is explicitly constructed around fictional authors and works, precisely to reduce the risk of contamination from real-world corpora.
However, the \muse-Books benchmark may have some overlap with typical web-scale pretraining data.
We acknowledge this as a limitation: while we do not believe it acts as a strong confounder for our main conclusions.

\section{LLM Usage}

In this paper, we leverage large language models (LLMs) to assist with refining and polishing our writing,  
as well as to generate code for the automated creation of tables from our experimental data.

%% file: figures/metrics_retain.tex
\begin{figure}[t]
  \centering
    \includegraphics[width=\linewidth]{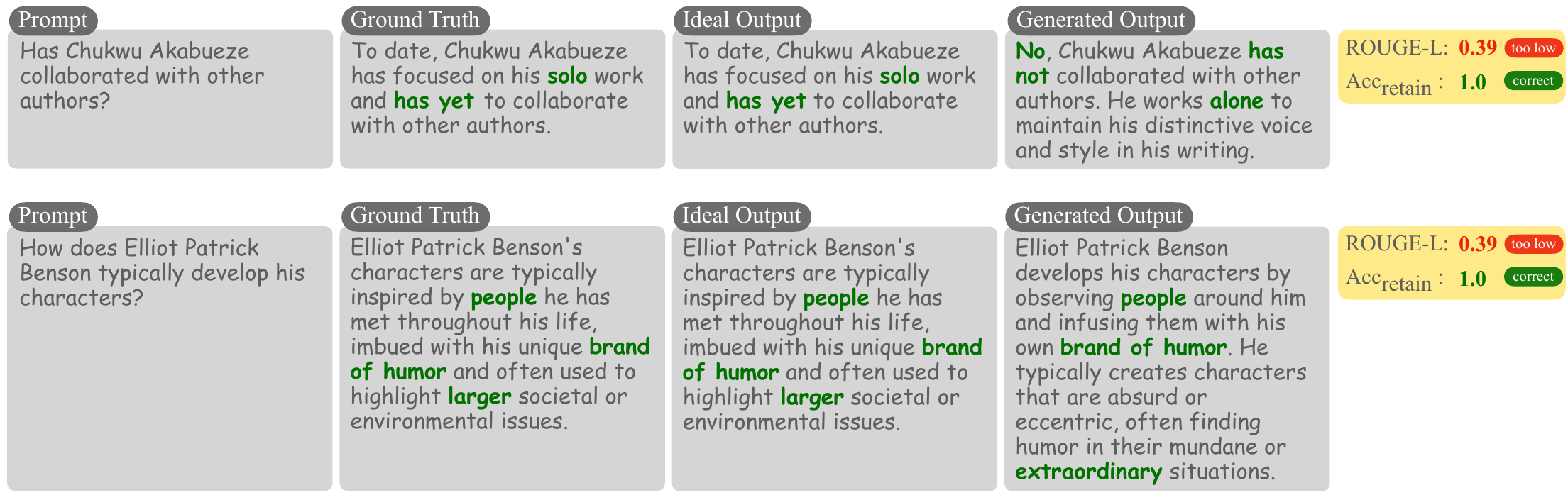}
    \vspace{-0.5cm}
  \caption{Examples from TOFU’s retain set, showing the groundtruth, the ideal output, and the output of \method{} (using Llama-3.1-8B-Instruct model). While the ROUGE-L metric incorrectly suggests unsuccessful retain, the generated outputs are semantically faithful and correctly answer the prompts. Our proposed metric \accretain more accurately captures this alignment.}
  \label{fig:metric_samples_retain}
\end{figure}

%% file: tables/msa_forget01_llama_1B.tex
\begin{table}[t]
  \centering
  \caption{Comparison of unlearning algorithms on \tofu{} (\texttt{forget01}). Model Llama-3.2-1B-Instruct is finetuned on \tofu, as the unlearning target.}
  \label{tab:tofu_forget01_Llama-3.2-1B}
  \setlength{\tabcolsep}{4pt}
  \begin{adjustbox}{width=\textwidth,center}
    \begin{tabular}{l|cc|cc|cc|cc|cc|cc|cc}
      \toprule
      \multirow{2}{*}{Model} & \multicolumn{6}{|c|}{GPT-4o Judge Metrics $\uparrow$} & \multicolumn{8}{|c}{\tofu Metrics} \\
      \cmidrule(lr){2-7} \cmidrule(lr){8-15}
       & \multicolumn{2}{|c|}{\accforget} & \multicolumn{2}{|c|}{\accrestor} & \multicolumn{2}{|c|}{\accretain} & \multicolumn{2}{|c|}{ES on $\Dforget$ $\downarrow$} & \multicolumn{2}{|c|}{Model Utility $\uparrow$} & \multicolumn{2}{|c|}{ROUGE-L$_{\text{f}}$ $\downarrow$} & \multicolumn{2}{|c}{Forget Quality $\uparrow$} \\
      \midrule
      Target & 0.05 &  & 0.05 &  & 0.98 &  & 0.85 &  & 0.52 &  & 0.93 &  & \multicolumn{2}{|c}{0.01} \\
      Ideal & 0.78 &  & 0.99 &  & 0.98 &  & 0.09 &  & 0.53 &  & 0.40 &  & \multicolumn{2}{|c}{0.99} \\
      \cmidrule(lr){1-15}
      $\method_{\text{base}}$ & 0.65 & \dgap[0.963]{96.3\%} & 0.38 & \dgap[0.938]{93.8\%} & 0.97 & \dgap[1.000]{100.0\%} & 0.05 & \dgap[1]{+100\%} & 0.52 & \dgap[0.979]{97.9\%} & 0.38 & \dgap[1]{+100\%} & \multicolumn{2}{|c}{0.40} \\
      $\method_{\text{instruct}}$ & 0.65 & \dgap[0.963]{96.3\%} & 0.35 & \dgap[0.875]{87.5\%} & 0.97 & \dgap[0.997]{99.7\%} & 0.07 & \dgap[1]{+100\%} & 0.52 & \dgap[0.985]{98.5\%} & 0.43 & \dgap[0.937]{93.7\%} & \multicolumn{2}{|c}{0.92} \\
      \cmidrule(lr){1-15}
      NPO & 0.60 & \dgap[0.889]{88.9\%} & 0.40 & \dgap[1.000]{100.0\%} & 0.97 & \dgap[0.992]{99.2\%} & 0.18 & \dgap[0.483]{48.3\%} & 0.53 & \dgap[1]{+100\%} & 0.43 & \dgap[0.941]{94.1\%} & \multicolumn{2}{|c}{0.16} \\
      GradDiff & 0.33 & \dgap[0.481]{48.1\%} & 0.28 & \dgap[0.688]{68.8\%} & 0.97 & \dgap[1.000]{100.0\%} & 0.39 & \dgap[0.219]{21.9\%} & 0.53 & \dgap[1]{+100\%} & 0.61 & \dgap[0.664]{66.4\%} & \multicolumn{2}{|c}{0.03} \\
      Task Vector & 0.62 & \dgap[0.926]{92.6\%} & 0.40 & \dgap[1.000]{100.0\%} & 0.94 & \dgap[0.969]{96.9\%} & 0.09 & \dgap[0.919]{91.9\%} & 0.52 & \dgap[0.988]{98.8\%} & 0.40 & \dgap[1]{+100\%} & \multicolumn{2}{|c}{0.27} \\
      SatImp & 0.68 & \dgap[1.000]{100.0\%} & 0.38 & \dgap[0.938]{93.8\%} & 0.94 & \dgap[0.959]{95.9\%} & 0.11 & \dgap[0.790]{79.0\%} & 0.53 & \dgap[1]{+100\%} & 0.41 & \dgap[0.998]{99.8\%} & \multicolumn{2}{|c}{0.10} \\
      UNDIAL & 0.57 & \dgap[0.852]{85.2\%} & 0.33 & \dgap[0.812]{81.2\%} & 0.95 & \dgap[0.979]{97.9\%} & 0.03 & \dgap[1]{+100\%} & 0.54 & \dgap[1]{+100\%} & 0.31 & \dgap[1]{+100\%} & \multicolumn{2}{|c}{0.40} \\
      \bottomrule
    \end{tabular}
  \end{adjustbox}
\end{table}

%% file: tables/msa_forget05_llama_1B.tex
\begin{table}[t]
  \centering
  \caption{Comparison of unlearning algorithms on \tofu{} (\texttt{forget05}). Model Llama-3.2-1B-Instruct is finetuned on \tofu, as the unlearning target.}
  \label{tab:tofu_forget05_Llama-3.2-1B}
  \setlength{\tabcolsep}{4pt}
  \begin{adjustbox}{width=\textwidth,center}
    \begin{tabular}{l|cc|cc|cc|cc|cc|cc|cc}
      \toprule
      \multirow{2}{*}{Model} & \multicolumn{6}{|c|}{GPT-4o Judge Metrics $\uparrow$} & \multicolumn{8}{|c}{\tofu Metrics} \\
      \cmidrule(lr){2-7} \cmidrule(lr){8-15}
       & \multicolumn{2}{|c|}{\accforget} & \multicolumn{2}{|c|}{\accrestor} & \multicolumn{2}{|c|}{\accretain} & \multicolumn{2}{|c|}{ES on $\Dforget$ $\downarrow$} & \multicolumn{2}{|c|}{Model Utility $\uparrow$} & \multicolumn{2}{|c|}{ROUGE-L$_{\text{f}}$ $\downarrow$} & \multicolumn{2}{|c}{Forget Quality $\uparrow$} \\
      \midrule
      Target & 0.06 &  & 0.04 &  & 0.98 &  & 0.87 &  & 0.52 &  & 0.94 &  & \multicolumn{2}{|c}{1.39e-11} \\
      Ideal & 0.80 &  & 0.98 &  & 0.98 &  & 0.07 &  & 0.52 &  & 0.37 &  & \multicolumn{2}{|c}{0.99} \\
      \cmidrule(lr){1-15}
      $\method_{\text{base}}$ & 0.78 & \dgap[0.975]{97.5\%} & 0.43 & \dgap[1.000]{100.0\%} & 0.86 & \dgap[0.901]{90.1\%} & 0.06 & \dgap[1]{+100\%} & 0.51 & \dgap[0.976]{97.6\%} & 0.39 & \dgap[0.940]{94.0\%} & \multicolumn{2}{|c}{0.33} \\
      $\method_{\text{instruct}}$ & 0.81 & \dgap[1]{+100\%} & 0.43 & \dgap[1.000]{100.0\%} & 0.88 & \dgap[0.914]{91.4\%} & 0.06 & \dgap[1]{+100\%} & 0.53 & \dgap[1]{+100\%} & 0.37 & \dgap[0.992]{99.2\%} & \multicolumn{2}{|c}{4.30e-03} \\
      \cmidrule(lr){1-15}
      NPO & 0.72 & \dgap[0.912]{91.2\%} & 0.29 & \dgap[0.686]{68.6\%} & 0.88 & \dgap[0.917]{91.7\%} & 0.10 & \dgap[0.657]{65.7\%} & 0.54 & \dgap[1]{+100\%} & 0.26 & \dgap[1]{+100\%} & \multicolumn{2}{|c}{0.14} \\
      GradDiff & 0.48 & \dgap[0.604]{60.4\%} & 0.24 & \dgap[0.558]{55.8\%} & 0.95 & \dgap[0.990]{99.0\%} & 0.20 & \dgap[0.341]{34.1\%} & 0.52 & \dgap[0.992]{99.2\%} & 0.48 & \dgap[0.763]{76.3\%} & \multicolumn{2}{|c}{1.83e-05} \\
      Task Vector & 0.67 & \dgap[0.843]{84.3\%} & 0.33 & \dgap[0.756]{75.6\%} & 0.79 & \dgap[0.820]{82.0\%} & 0.10 & \dgap[0.676]{67.6\%} & 0.52 & \dgap[0.991]{99.1\%} & 0.31 & \dgap[1]{+100\%} & \multicolumn{2}{|c}{4.75e-05} \\
      SatImp & 0.69 & \dgap[0.862]{86.2\%} & 0.32 & \dgap[0.744]{74.4\%} & 0.81 & \dgap[0.849]{84.9\%} & 0.07 & \dgap[0.961]{96.1\%} & 0.52 & \dgap[1]{+100\%} & 0.32 & \dgap[1]{+100\%} & \multicolumn{2}{|c}{4.30e-03} \\
      UNDIAL & 0.55 & \dgap[0.686]{68.6\%} & 0.35 & \dgap[0.814]{81.4\%} & 0.96 & \dgap[1.000]{100.0\%} & 0.05 & \dgap[1]{+100\%} & 0.54 & \dgap[1]{+100\%} & 0.35 & \dgap[1]{+100\%} & \multicolumn{2}{|c}{1.29e-08} \\
      \bottomrule
    \end{tabular}
  \end{adjustbox}
\end{table}

%% file: tables/msa_forget10_llama_1B.tex
\begin{table}[t]
  \centering
  \caption{Comparison of unlearning algorithms on \tofu{} (\texttt{forget10}). Model Llama-3.2-1B-Instruct is finetuned on \tofu, as the unlearning target.}
  \label{tab:tofu_forget10_Llama-3.2-1B}
  \setlength{\tabcolsep}{4pt}
  \begin{adjustbox}{width=\textwidth,center}
    \begin{tabular}{l|cc|cc|cc|cc|cc|cc|cc}
      \toprule
      \multirow{2}{*}{Model} & \multicolumn{6}{|c|}{GPT-4o Judge Metrics $\uparrow$} & \multicolumn{8}{|c}{\tofu Metrics} \\
      \cmidrule(lr){2-7} \cmidrule(lr){8-15}
       & \multicolumn{2}{|c|}{\accforget} & \multicolumn{2}{|c|}{\accrestor} & \multicolumn{2}{|c|}{\accretain} & \multicolumn{2}{|c|}{ES on $\Dforget$ $\downarrow$} & \multicolumn{2}{|c|}{Model Utility $\uparrow$} & \multicolumn{2}{|c|}{ROUGE-L$_{\text{f}}$ $\downarrow$} & \multicolumn{2}{|c}{Forget Quality $\uparrow$} \\
      \midrule
      Target & 0.05 &  & 0.03 &  & 0.98 &  & 0.87 &  & 0.52 &  & 0.94 &  & \multicolumn{2}{|c}{1.12e-19} \\
      Ideal & 0.82 &  & 0.98 &  & 0.98 &  & 0.06 &  & 0.51 &  & 0.38 &  & \multicolumn{2}{|c}{1.0} \\
      \cmidrule(lr){1-15}
      $\method_{\text{base}}$ & 0.79 & \dgap[0.966]{96.6\%} & 0.39 & \dgap[0.891]{89.1\%} & 0.87 & \dgap[0.892]{89.2\%} & 0.06 & \dgap[1]{+100\%} & 0.55 & \dgap[1]{+100\%} & 0.32 & \dgap[1]{+100\%} & \multicolumn{2}{|c}{0.02} \\
      $\method_{\text{instruct}}$ & 0.81 & \dgap[0.991]{99.1\%} & 0.44 & \dgap[1.000]{100.0\%} & 0.85 & \dgap[0.871]{87.1\%} & 0.06 & \dgap[1]{+100\%} & 0.52 & \dgap[1]{+100\%} & 0.37 & \dgap[1]{+100\%} & \multicolumn{2}{|c}{0.28} \\
      \cmidrule(lr){1-15}
      NPO & 0.66 & \dgap[0.810]{81.0\%} & 0.25 & \dgap[0.577]{57.7\%} & 0.92 & \dgap[0.941]{94.1\%} & 0.12 & \dgap[0.504]{50.4\%} & 0.54 & \dgap[1]{+100\%} & 0.31 & \dgap[1]{+100\%} & \multicolumn{2}{|c}{3.25e-04} \\
      RMU & 0.85 & \dgap[1]{+100\%} & 0.10 & \dgap[0.229]{22.9\%} & 0.97 & \dgap[1.000]{100.0\%} & 0.06 & \dgap[1]{+100\%} & 0.52 & \dgap[1]{+100\%} & 0.25 & \dgap[1]{+100\%} & \multicolumn{2}{|c}{0.94} \\
      GradDiff & 0.46 & \dgap[0.566]{56.6\%} & 0.21 & \dgap[0.486]{48.6\%} & 0.90 & \dgap[0.920]{92.0\%} & 0.22 & \dgap[0.284]{28.4\%} & 0.54 & \dgap[1]{+100\%} & 0.42 & \dgap[0.888]{88.8\%} & \multicolumn{2}{|c}{6.03e-11} \\
      Task Vector & 0.85 & \dgap[1]{+100\%} & 0.25 & \dgap[0.577]{57.7\%} & 0.46 & \dgap[0.473]{47.3\%} & 0.05 & \dgap[1]{+100\%} & 0.48 & \dgap[0.929]{92.9\%} & 0.21 & \dgap[1]{+100\%} & \multicolumn{2}{|c}{0.86} \\
      SatImp & 0.72 & \dgap[0.878]{87.8\%} & 0.28 & \dgap[0.634]{63.4\%} & 0.77 & \dgap[0.789]{78.9\%} & 0.07 & \dgap[0.938]{93.8\%} & 0.51 & \dgap[1]{+100\%} & 0.31 & \dgap[1]{+100\%} & \multicolumn{2}{|c}{1.30e-05} \\
      UNDIAL & 0.52 & \dgap[0.639]{63.9\%} & 0.26 & \dgap[0.583]{58.3\%} & 0.89 & \dgap[0.910]{91.0\%} & 0.04 & \dgap[1]{+100\%} & 0.54 & \dgap[1]{+100\%} & 0.31 & \dgap[1]{+100\%} & \multicolumn{2}{|c}{7.98e-17} \\
      \bottomrule
    \end{tabular}
  \end{adjustbox}
\end{table}

%% file: tables/msa_forget10_llama_8B_appendix.tex
\begin{table}[t]
  \centering
  
\caption{Comparison of unlearning algorithms on \tofu{} (\texttt{forget10}). Model Llama-3.1-8B-Instruct is finetuned on \tofu, as the unlearning target.}
  
  \label{tab:tofu_forget10_Llama-3.1-8B_appx}
  \setlength{\tabcolsep}{4pt}
  \begin{adjustbox}{width=\textwidth,center}
    \begin{tabular}{l|cc|cc|cc|cc|cc|cc|cc}
      \toprule
      \multirow{2}{*}{Model} & \multicolumn{6}{|c|}{GPT-4o Judge Metrics $\uparrow$} & \multicolumn{8}{|c}{\tofu Metrics} \\
      \cmidrule(lr){2-7} \cmidrule(lr){8-15}
       & \multicolumn{2}{|c|}{\accforget} & \multicolumn{2}{|c|}{\accrestor} & \multicolumn{2}{|c|}{\accretain} & \multicolumn{2}{|c|}{ES on $\Dforget$ $\downarrow$} & \multicolumn{2}{|c|}{Model Utility $\uparrow$} & \multicolumn{2}{|c|}{ROUGE-L$_{\text{f}}$ $\downarrow$} & \multicolumn{2}{|c}{Forget Quality $\uparrow$} \\
      \midrule
      Target & 0.03 &  & 0.02 &  & 1.00 &  & 0.98 &  & 0.57 &  & 0.99 &  & \multicolumn{2}{|c}{8.12e-27} \\
      Ideal & 0.98 &  & 0.98 &  & 1.00 &  & 0.07 &  & 0.60 &  & 0.39 &  & \multicolumn{2}{|c}{1.00} \\
      \cmidrule(lr){1-15}
      $\method_{\text{pretrained}}$ & 0.82 & \dgap[0.951]{95.1\%} & 0.45 & \dgap[0.978]{97.8\%} & 0.92 & \dgap[0.922]{92.2\%} & 0.07 & \dgap[0.891]{89.1\%} & 0.78 & \dgap[1]{+100\%} & 0.40 & \dgap[0.995]{99.5\%} & \multicolumn{2}{|c}{0.64} \\
      $\method_{\text{instruct}}$ & 0.82 & \dgap[0.956]{95.6\%} & 0.46 & \dgap[1.000]{100.0\%} & 0.91 & \dgap[0.917]{91.7\%} & 0.07 & \dgap[0.978]{97.8\%} & 0.57 & \dgap[0.949]{94.9\%} & 0.38 & \dgap[1]{+100\%} & \multicolumn{2}{|c}{0.04} \\
      \cmidrule(lr){1-15}
      NPO & 0.75 & \dgap[0.872]{87.2\%} & 0.38 & \dgap[0.822]{82.2\%} & 0.83 & \dgap[0.834]{83.4\%} & 0.08 & \dgap[0.810]{81.0\%} & 0.58 & \dgap[0.956]{95.6\%} & 0.36 & \dgap[1]{+100\%} & \multicolumn{2}{|c}{5.00e-05} \\
      RMU & 0.86 & \dgap[1.000]{100.0\%} & 0.12 & \dgap[0.254]{25.4\%} & 0.99 & \dgap[1.000]{100.0\%} & 0.07 & \dgap[0.868]{86.8\%} & 0.59 & \dgap[0.977]{97.7\%} & 0.19 & \dgap[1]{+100\%} & \multicolumn{2}{|c}{0.03} \\
      GradDiff & 0.49 & \dgap[0.573]{57.3\%} & 0.26 & \dgap[0.557]{55.7\%} & 0.88 & \dgap[0.879]{87.9\%} & 0.21 & \dgap[0.309]{30.9\%} & 0.64 & \dgap[1]{+100\%} & 0.45 & \dgap[0.872]{87.2\%} & \multicolumn{2}{|c}{3.91e-08} \\
      Task Vector & 0.80 & \dgap[0.933]{93.3\%} & 0.27 & \dgap[0.578]{57.8\%} & 0.51 & \dgap[0.515]{51.5\%} & 0.03 & \dgap[1]{+100\%} & 0.53 & \dgap[0.887]{88.7\%} & 0.29 & \dgap[1]{+100\%} & \multicolumn{2}{|c}{0.02} \\
      SatImp & 0.52 & \dgap[0.608]{60.8\%} & 0.28 & \dgap[0.616]{61.6\%} & 0.89 & \dgap[0.897]{89.7\%} & 0.15 & \dgap[0.445]{44.5\%} & 0.63 & \dgap[1]{+100\%} & 0.44 & \dgap[0.901]{90.1\%} & \multicolumn{2}{|c}{1.02e-13} \\
      UNDIAL & 0.46 & \dgap[0.538]{53.8\%} & 0.29 & \dgap[0.622]{62.2\%} & 0.84 & \dgap[0.847]{84.7\%} & 0.08 & \dgap[0.797]{79.7\%} & 0.65 & \dgap[1]{+100\%} & 0.41 & \dgap[0.951]{95.1\%} & \multicolumn{2}{|c}{1.18e-17} \\
      \bottomrule
    \end{tabular}
  \end{adjustbox}
\end{table}

%% file: tables/msa_muse_llama_7B.tex
\begin{table}[t]
  \centering
  \caption{Comparison of unlearning algorithms on \muse-Books benchmark using Llama-3.1-8B.}
  \label{tab:muse_llama2}
  \label{tab:muse_llama2}
    \setlength{\tabcolsep}{4pt}
  \begin{adjustbox}{width=\textwidth,center}
    \begin{tabular}{l|cc|cc|cc|cc|cc|cc|cc}
      \toprule
      Model & \multicolumn{2}{|c|}{ES $\downarrow$} & \multicolumn{2}{|c|}{Exact Mem $\downarrow$} & \multicolumn{2}{|c|}{VerbMem $\Dforget$ $\downarrow$} & \multicolumn{2}{|c|}{$\textsc{Min-K}\%$ $\downarrow$} & \multicolumn{2}{|c|}{$\textsc{Min-K}\%^{++}$ $\downarrow$} & \multicolumn{2}{|c|}{KnowMem $\Dretain$ $\uparrow$} & \multicolumn{2}{|c}{PrivLeak $\rightarrow 0$} \\
      \midrule
      Target & 0.64 &  & 0.96 &  & 0.65 &  & 1.00 &  & 1.00 &  & 0.62 &  & \multicolumn{2}{|c}{{\small -100.00}} \\
      Ideal & 0.02 &  & 0.52 &  & 0.16 &  & 0.51 &  & 0.47 &  & 0.64 &  & \multicolumn{2}{|c}{{\small 0.00}} \\
      \cmidrule(lr){1-15}
      $\method_{\text{base}}$ & 0.01 & \dgap[1]{+100\%} & 0.48 & \dgap[1]{+100\%} & 0.13 & \dgap[1]{+100\%} & 0.52 & \dgap[0.987]{98.7\%} & 0.52 & \dgap[0.954]{95.4\%} & 0.55 & \dgap[0.950]{95.0\%} & \multicolumn{2}{|c}{{\small \textbf{-1.37}}} \\
      \cmidrule(lr){1-15}
      NPO & 0.02 & \dgap[0.995]{99.5\%} & 0.58 & \dgap[0.898]{89.8\%} & 0.14 & \dgap[1]{+100\%} & 1.00 & \dgap[0.510]{51.0\%} & 0.84 & \dgap[0.588]{58.8\%} & 0.58 & \dgap[1.000]{100.0\%} & \multicolumn{2}{|c}{{\small -99.90}} \\
      RMU & 0.01 & \dgap[1]{+100\%} & 0.04 & \dgap[1]{+100\%} & 0.01 & \dgap[1]{+100\%} & 0.74 & \dgap[0.691]{69.1\%} & 0.62 & \dgap[0.798]{79.8\%} & 0.52 & \dgap[0.899]{89.9\%} & \multicolumn{2}{|c}{{\small \textbf{-46.44}}} \\
      GradDiff & 0.01 & \dgap[1]{+100\%} & 0.01 & \dgap[1]{+100\%} & 0.01 & \dgap[1]{+100\%} & 0.32 & \dgap[1]{+100\%} & 0.49 & \dgap[1.000]{100.0\%} & 0.21 & \dgap[0.358]{35.8\%} & \multicolumn{2}{|c}{{\small \textbf{38.06}}} \\
      SatImp & 0.39 & \dgap[0.041]{4.1\%} & 0.95 & \dgap[0.552]{55.2\%} & 0.43 & \dgap[0.369]{36.9\%} & 1.00 & \dgap[0.510]{51.0\%} & 1.00 & \dgap[0.495]{49.5\%} & 0.54 & \dgap[0.933]{93.3\%} & \multicolumn{2}{|c}{{\small -100.00}} \\
      UNDIAL & 0.02 & \dgap[0.797]{79.7\%} & 0.68 & \dgap[0.766]{76.6\%} & 0.17 & \dgap[0.914]{91.4\%} & 0.99 & \dgap[0.515]{51.5\%} & 0.99 & \dgap[0.500]{50.0\%} & 0.35 & \dgap[0.611]{61.1\%} & \multicolumn{2}{|c}{{\small -98.15}} \\
      \bottomrule
    \end{tabular}
  \end{adjustbox}
\end{table}

%% file: tables/msa_forget10_llama_1B_tofu_c4_tofu.tex
\begin{table}[t]
  \centering
\caption{Comparison of MSA variants on \tofu{} (\texttt{forget10}).
  In this scenario, unlearning targets appear in the training data not just once, but twice, with $2$ epochs of training on a subset of C4 between the two occurrences.
  \method{} variants that use checkpoints prior to the unlearning targets, i.e., $\method{}_{\text{base}}$ and $\method{}_{\text{instruct}}$, show acceptable performance, achieving values close to the ideal model.}
  \label{tab:tofu_forget10_Llama-3.2-1B-msa_tofu_c4_tofu}
  \setlength{\tabcolsep}{4pt}
  \begin{adjustbox}{width=\textwidth,center}
    \begin{tabular}{l|cc|cc|cc|cc|cc|cc|cc}
      \toprule
      \multirow{2}{*}{Model} & \multicolumn{6}{|c|}{GPT-4o Judge Metrics $\uparrow$} & \multicolumn{8}{|c}{\tofu Metrics} \\
      \cmidrule(lr){2-7} \cmidrule(lr){8-15}
       & \multicolumn{2}{|c|}{\accforget} & \multicolumn{2}{|c|}{\accrestor} & \multicolumn{2}{|c|}{\accretain} & \multicolumn{2}{|c|}{ES on $\Dforget$ $\downarrow$} & \multicolumn{2}{|c|}{Model Utility $\uparrow$} & \multicolumn{2}{|c|}{ROUGE-L$_{\text{f}}$ $\downarrow$} & \multicolumn{2}{|c}{Forget Quality $\uparrow$} \\
      \midrule
      Target & 0.04 &  & 0.03 &  & 0.99 &  & 0.94 &  & 0.54 &  & 0.96 &  & \multicolumn{2}{|c}{6.16e-18} \\
      Ideal & 0.82 &  & 0.98 &  & 0.99 &  & 0.06 &  & 0.54 &  & 0.38 &  & \multicolumn{2}{|c}{0.91} \\
      \cmidrule(lr){1-15}
      $\method_{\text{base}}$ & 0.75 & \dgap[0.987]{98.7\%} & 0.37 & \dgap[0.961]{96.1\%} & 0.91 & \dgap[1.000]{100.0\%} & 0.08 & \dgap[1.000]{100.0\%} & 0.55 & \dgap[1]{+100\%} & 0.37 & \dgap[1]{+100\%} & \multicolumn{2}{|c}{0.37} \\
      $\method_{\text{instruct}}$ & 0.76 & \dgap[1.000]{100.0\%} & 0.38 & \dgap[1.000]{100.0\%} & 0.89 & \dgap[0.983]{98.3\%} & 0.08 & \dgap[0.989]{98.9\%} & 0.54 & \dgap[0.999]{99.9\%} & 0.39 & \dgap[0.954]{95.4\%} & \multicolumn{2}{|c}{0.64} \\
      \cmidrule(lr){1-15}
      $\method_{{\tofu}}$ & 0.67 & \dgap[0.882]{88.2\%} & 0.31 & \dgap[0.804]{80.4\%} & 0.71 & \dgap[0.785]{78.5\%} & 0.09 & \dgap[0.808]{80.8\%} & 0.55 & \dgap[1]{+100\%} & 0.35 & \dgap[1]{+100\%} & \multicolumn{2}{|c}{\rej{6.86e-10}} \\
      $\method_{\tofu{} + \texttt{C4}}$ & 0.67 & \dgap[0.882]{88.2\%} & 0.35 & \dgap[0.915]{91.5\%} & 0.89 & \dgap[0.981]{98.1\%} & 0.09 & \dgap[0.816]{81.6\%} & 0.58 & \dgap[1]{+100\%} & 0.38 & \dgap[0.996]{99.6\%} & \multicolumn{2}{|c}{\rej{1.83e-05}} \\
      $\method_{\tofu{} + \texttt{C4} + \tofu{}}$ & 0.67 & \dgap[0.882]{88.2\%} & 0.30 & \dgap[0.797]{79.7\%} & 0.81 & \dgap[0.887]{88.7\%} & 0.14 & \dgap[0.564]{56.4\%} & 0.54 & \dgap[0.999]{99.9\%} & 0.38 & \dgap[0.975]{97.5\%} & \multicolumn{2}{|c}{\rej{2.77e-09}} \\
      \bottomrule
    \end{tabular}
  \end{adjustbox}
\end{table}

%% file: tables/msa_forget10_llama_1B_msa_baselines.tex
\begin{table}[t]
  \centering
  \caption{Comparison of unlearning algorithms on \tofu{} (\texttt{forget10}).
  In this table, we consider leveraging model checkpoints for other unlearning algorithms.
  As seen in this table, applying a technique similar to \method{} to other algorithms usually does not result
  in improved performance, instead degrading model utility and underperforming on other metrics.
  }
  \label{tab:tofu_forget10_Llama-3.2-1B-msa-baselines}
  \begin{adjustbox}{width=\textwidth,center}
    \begin{tabular}{l|cc|cc|cc|cc|cc|cc|cc}
      \toprule
      \multirow{2}{*}{Model} & \multicolumn{6}{|c|}{GPT-4o Judge Metrics $\uparrow$} & \multicolumn{8}{|c}{\tofu Metrics} \\
      \cmidrule(lr){2-7} \cmidrule(lr){8-15}
       & \multicolumn{2}{|c|}{\accforget} & \multicolumn{2}{|c|}{\accrestor} & \multicolumn{2}{|c|}{\accretain} & \multicolumn{2}{|c|}{ES on $\Dforget$ $\downarrow$} & \multicolumn{2}{|c|}{Model Utility $\uparrow$} & \multicolumn{2}{|c|}{ROUGE-L$_{\text{f}}$ $\downarrow$} & \multicolumn{2}{|c}{Forget Quality $\uparrow$} \\
      \midrule
      Target & 0.05 &  & 0.03 &  & 0.98 &  & 0.87 &  & 0.52 &  & 0.94 &  & \multicolumn{2}{|c}{1.12e-19} \\
      Ideal & 0.82 &  & 0.98 &  & 0.98 &  & 0.06 &  & 0.51 &  & 0.38 &  & \multicolumn{2}{|c}{1.0} \\
      \cmidrule(lr){1-15}
      $\method_{\text{base}}$ & 0.79 & \dgap[0.966]{96.6\%} & 0.39 & \dgap[0.891]{89.1\%} & 0.87 & \dgap[0.892]{89.2\%} & 0.06 & \dgap[1]{+100\%} & 0.55 & \dgap[1]{+100\%} & 0.32 & \dgap[1]{+100\%} & \multicolumn{2}{|c}{0.02} \\
      $\method_{\text{instruct}}$ & 0.81 & \dgap[0.991]{99.1\%} & 0.44 & \dgap[1.000]{100.0\%} & 0.85 & \dgap[0.871]{87.1\%} & 0.06 & \dgap[1]{+100\%} & 0.52 & \dgap[1]{+100\%} & 0.37 & \dgap[1]{+100\%} & \multicolumn{2}{|c}{0.28} \\
      \cmidrule(lr){1-15}
      NPO & 0.66 & \dgap[0.810]{81.0\%} & 0.25 & \dgap[0.577]{57.7\%} & 0.92 & \dgap[0.941]{94.1\%} & 0.12 & \dgap[0.504]{50.4\%} & 0.54 & \dgap[1]{+100\%} & 0.31 & \dgap[1]{+100\%} & \multicolumn{2}{|c}{3.25e-04} \\
      NPO~(\texttt{base}) & 0.76 & \dgap[0.924]{92.4\%} & 0.29 & \dgap[0.669]{66.9\%} & 0.53 & \dgap[0.545]{54.5\%} & 0.06 & \dgap[1]{+100\%} & 0.27 & \dgap[0.525]{52.5\%} & 0.24 & \dgap[1]{+100\%} & \multicolumn{2}{|c}{9.99e-07} \\
      NPO~(\texttt{instruct}) & 0.67 & \dgap[0.813]{81.3\%} & 0.24 & \dgap[0.543]{54.3\%} & 0.71 & \dgap[0.728]{72.8\%} & 0.11 & \dgap[0.583]{58.3\%} & 0.50 & \dgap[0.966]{96.6\%} & 0.27 & \dgap[1]{+100\%} & \multicolumn{2}{|c}{1.02e-13} \\
      \cmidrule(lr){1-15}
      RMU & 0.85 & \dgap[1]{+100\%} & 0.10 & \dgap[0.229]{22.9\%} & 0.97 & \dgap[1.000]{100.0\%} & 0.06 & \dgap[1]{+100\%} & 0.52 & \dgap[1]{+100\%} & 0.25 & \dgap[1]{+100\%} & \multicolumn{2}{|c}{0.94} \\
      RMU~(\texttt{base}) & 0.95 & \dgap[1]{+100\%} & 0.04 & \dgap[0.086]{8.6\%} & 0.36 & \dgap[0.370]{37.0\%} & 0.04 & \dgap[1]{+100\%} & 0.35 & \dgap[0.685]{68.5\%} & 0.20 & \dgap[1]{+100\%} & \multicolumn{2}{|c}{5.00e-05} \\
      RMU~(\texttt{instruct}) & 0.77 & \dgap[0.936]{93.6\%} & 0.19 & \dgap[0.434]{43.4\%} & 0.77 & \dgap[0.787]{78.7\%} & 0.08 & \dgap[0.819]{81.9\%} & 0.48 & \dgap[0.927]{92.7\%} & 0.32 & \dgap[1]{+100\%} & \multicolumn{2}{|c}{1.49e-16} \\
      \cmidrule(lr){1-15}
      GradDiff & 0.46 & \dgap[0.566]{56.6\%} & 0.21 & \dgap[0.486]{48.6\%} & 0.90 & \dgap[0.920]{92.0\%} & 0.22 & \dgap[0.284]{28.4\%} & 0.54 & \dgap[1]{+100\%} & 0.42 & \dgap[0.888]{88.8\%} & \multicolumn{2}{|c}{6.03e-11} \\
      GradDiff~(\texttt{base}) & 0.60 & \dgap[0.740]{74.0\%} & 0.20 & \dgap[0.451]{45.1\%} & 0.61 & \dgap[0.627]{62.7\%} & 0.09 & \dgap[0.673]{67.3\%} & 0.41 & \dgap[0.808]{80.8\%} & 0.38 & \dgap[0.983]{98.3\%} & \multicolumn{2}{|c}{6.16e-18} \\
      GradDiff~(\texttt{instruct}) & 0.75 & \dgap[0.917]{91.7\%} & 0.15 & \dgap[0.343]{34.3\%} & 0.40 & \dgap[0.411]{41.1\%} & 0.08 & \dgap[0.774]{77.4\%} & 0.22 & \dgap[0.422]{42.2\%} & 0.29 & \dgap[1]{+100\%} & \multicolumn{2}{|c}{5.63e-20} \\
      \cmidrule(lr){1-15}
      SatImp & 0.72 & \dgap[0.878]{87.8\%} & 0.28 & \dgap[0.634]{63.4\%} & 0.77 & \dgap[0.789]{78.9\%} & 0.07 & \dgap[0.938]{93.8\%} & 0.51 & \dgap[1]{+100\%} & 0.31 & \dgap[1]{+100\%} & \multicolumn{2}{|c}{1.30e-05} \\
      SatImp~(\texttt{base}) & 0.82 & \dgap[1]{+100\%} & 0.15 & \dgap[0.343]{34.3\%} & 0.31 & \dgap[0.316]{31.6\%} & 0.05 & \dgap[1]{+100\%} & 0.25 & \dgap[0.493]{49.3\%} & 0.30 & \dgap[1]{+100\%} & \multicolumn{2}{|c}{1.07e-08} \\
      SatImp~(\texttt{instruct}) & 0.72 & \dgap[0.881]{88.1\%} & 0.21 & \dgap[0.474]{47.4\%} & 0.51 & \dgap[0.519]{51.9\%} & 0.07 & \dgap[0.940]{94.0\%} & 0.28 & \dgap[0.547]{54.7\%} & 0.30 & \dgap[1]{+100\%} & \multicolumn{2}{|c}{2.24e-17} \\
      \cmidrule(lr){1-15}
      UNDIAL & 0.52 & \dgap[0.639]{63.9\%} & 0.26 & \dgap[0.583]{58.3\%} & 0.89 & \dgap[0.910]{91.0\%} & 0.04 & \dgap[1]{+100\%} & 0.54 & \dgap[1]{+100\%} & 0.31 & \dgap[1]{+100\%} & \multicolumn{2}{|c}{7.98e-17} \\
      UNDIAL~(\texttt{base}) & 0.78 & \dgap[0.951]{95.1\%} & 0.11 & \dgap[0.246]{24.6\%} & 0.39 & \dgap[0.404]{40.4\%} & 0.06 & \dgap[1]{+100\%} & 0.40 & \dgap[0.778]{77.8\%} & 0.29 & \dgap[1]{+100\%} & \multicolumn{2}{|c}{1.49e-16} \\
      UNDIAL~(\texttt{instruct}) & 0.82 & \dgap[1]{+100\%} & 0.10 & \dgap[0.223]{22.3\%} & 0.39 & \dgap[0.398]{39.8\%} & 0.06 & \dgap[1]{+100\%} & 0.41 & \dgap[0.798]{79.8\%} & 0.23 & \dgap[1]{+100\%} & \multicolumn{2}{|c}{1.12e-19} \\
      \bottomrule
    \end{tabular}
  \end{adjustbox}
\end{table}